\documentclass[10pt,twocolumn,letterpaper]{article}

\usepackage{cvpr}              

\usepackage[dvipsnames]{xcolor}

\usepackage{graphicx}

\usepackage{graphicx}
\usepackage{amsmath}
\usepackage{amssymb}
\usepackage{multicol, blindtext}
\usepackage{subcaption}
\usepackage{cite}
\usepackage{soul}
\usepackage{units}
\usepackage{array}
\usepackage{multirow}
\usepackage{caption}
\usepackage{float}
\usepackage{booktabs}
\usepackage{times}
\usepackage{tabularx}
\usepackage{nicefrac}
\usepackage{enumitem}
\usepackage[nolist]{acronym}

\usepackage{adjustbox}

\newcolumntype{R}[2]{
    >{\adjustbox{angle=#1,lap=\width-(#2)}\bgroup}%
    l
    <{\egroup}%
}

\newcolumntype{L}[1]{>{\raggedright\arraybackslash}p{#1}}
\newcolumntype{C}[1]{>{\centering\arraybackslash}p{#1}}
\newcolumntype{R}[1]{>{\raggedleft\arraybackslash}p{#1}}
\usepackage{pifont}
\newcommand{\cmark}{\ding{51}}
\newcommand{\xmark}{\ding{55}}

\usepackage{caption}
\usepackage{xcolor}
\usepackage{mathtools}
\usepackage{xfrac}
\usepackage{wasysym}
\usepackage{algorithm}%
\usepackage{algpseudocode}%
\usepackage[accsupp]{axessibility}  

\makeatletter

\@namedef{ver@everyshi.sty}{}
\newcommand\notsotiny{\@setfontsize\notsotiny{6.31415}{7.1828}} 

\makeatother

\usepackage{comment}
\usepackage{pgfplots}
\usepackage{tikz} 
\usetikzlibrary{patterns}
\pgfplotsset{compat=newest}

\newcommand{\PAR}[1]{\vspace{-0.2eM}\vskip4pt \noindent{\bf #1}}

\definecolor{cvprblue}{rgb}{0.21,0.49,0.74}
\usepackage[pagebackref,breaklinks,colorlinks,citecolor=cvprblue]{hyperref}

\def\paperID{11292} 
\def\confName{CVPR}
\def\confYear{2024}

\title{\vspace{-30pt}Gated Fields: Learning Scene Reconstruction from Gated Videos }
\author{\centerline{Andrea Ramazzina$^{1}$\thanks{These authors contributed equally to this work.}\quad Stefanie Walz$^{1,2*}$\quad Pragyan Dahal$^{3}$\quad Mario Bijelic$^{4,5}$ \quad  Felix Heide$^{4,5}$} \\
{\small\centerline{$^1$Mercedes-Benz  \quad $^2$Saarland University \quad $^3$Politecnico di Milano \quad $^4$Torc Robotics  \quad $^5$Princeton University}}
}

\begin{document}

\begin{acronym}
\acro{tof}[ToF]{Time-of-Flight}
\acro{nir}[NIR]{near-infrared}
\end{acronym}

\maketitle
\begin{abstract}
Reconstructing outdoor 3D scenes from temporal observations is a challenge that recent work on neural fields has offered a new avenue for. However, existing methods that recover scene properties, such as geometry, appearance, or radiance, solely from RGB captures often fail when handling poorly-lit or texture-deficient regions. Similarly, recovering scenes with scanning LiDAR sensors is also difficult due to their low angular sampling rate which makes recovering expansive real-world scenes difficult. Tackling these gaps, we introduce Gated Fields – a neural scene reconstruction method that utilizes active gated video sequences. To this end, we propose a neural rendering approach that seamlessly incorporates time-gated capture and illumination. Our method exploits the intrinsic depth cues in the gated videos, achieving precise and dense geometry reconstruction irrespective of ambient illumination conditions. We validate the method across day and night scenarios and find that Gated Fields compares favorably to RGB and LiDAR reconstruction methods. Our code and datasets are available \href{ https://light.princeton.edu/gatedfields/}{here}\footnote{\scriptsize\url{ https://light.princeton.edu/gatedfields/}\label{link}}.
\end{abstract}    
\section{Introduction}
\label{sec:intro}
Large-scale outdoor scene reconstruction is essential for advancing autonomous robotics, drones, and driver-assistance systems, serving as the foundation for scene understanding, safe navigation, dataset generation and validation.
Existing works in this domain \cite{yang2017real,pradeep2013monofusion,schops20153d} have typically adopted a two-step approach. Initially, they infer depth maps from different poses, utilizing time-of-flight sensors or RGB captures. Subsequently, these depth estimates are fused to produce a coherent 3D representation, using either classical methodologies \cite{curless1996tsdf} or learned-based representations \cite{riegler2017octnetfusion}. In contrast, more recent studies \cite{murez2020atlas,bozic2021transformerfusion,sun2021neuralrecon} have proposed end-to-end strategies that bypass the estimation of local depth maps as an intermediate representation. Instead, they directly regress a Truncated Signed Distance Function (TSDF) \cite{murez2020atlas,sun2021neuralrecon} or an occupancy volume \cite{bozic2021transformerfusion}.
A rapidly growing body of work on neural rendering \cite{mildenhall2020nerf,wang2021neus} offers not only geometrically-accurate scene reconstruction from posed RGB images but also to generate novel perspectives from unobserved angles. Hinging on implicit coordinate-based neural representations, RGB-based methods ~\cite{zhang2020nerf++,barron2022mipnerf360,barron2023zipnerf,li2023neuralangelo} have been adapted to large open outdoor environments. A recent line of work~\cite{rematas2022urf,turki2023suds,wang2023fegr, guo2023streetsurf} includes LiDAR scans for auxiliary depth supervision and to improve scene reconstruction for urban environments. However, recovery based on the RGB images exhibit is fundamentally limited in the presence of low light~\cite{mildenhall2022nerfdark,wang2023lightingnerf} or in the presence of scattering such as fog \cite{levy2023seathru,ramazzina2023scatternerf}. 

 \begin{figure}[t!]
	\centering
	\includegraphics[width=0.48\textwidth]{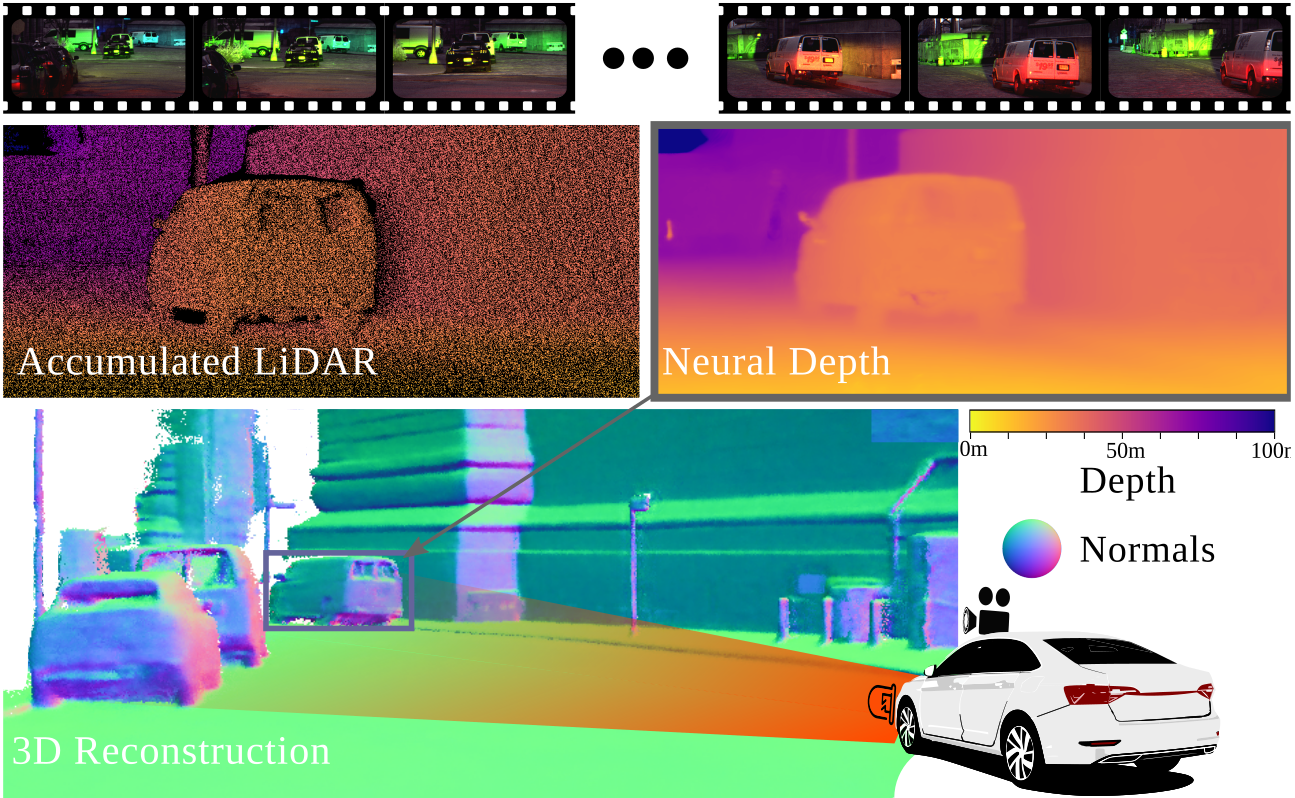}\vspace*{-2mm}
	\caption{From a single video of gated captures (top-row), we reconstruct an accurate scene representation and render depth projections (mid-row, right) as accurate as LiDAR scans (mid-row, left), and we recover 3D geometry and normals (bottom-row).
 }
	\label{fig:teaser}
	\vspace*{-5mm}
\end{figure}

A parallel direction of research investigates neural rendering techniques tailored to Time-of-Flight (ToF) sensors as opposed to the conventional RGB cameras. Existing methods \cite{huang2023nfl,tao2023lidarnerf} tackle scene understanding with posed LiDAR scans, and have modeled the raw output from a single-photon LiDAR system~\cite{malik2023transientnerf} or continuous-wave ToF sensor \cite{attal2021torf} as additional depth supervision. However, all of these existing methods struggle with recovering large unbounded outdoor scenes: while continuous-wave ToF sensors~\cite{attal2021torf} offer signal only in room-sized scenes, methods based on scanning LiDAR suffer from low angular sampling which mandates temporal aggregation. Specifically, even today's LiDAR sensors boasting 200 scan lines, lag drastically in resolution compared to current HDR cameras that offer two orders of magnitude higher vertical pixel counts nearing 10k and three orders of magnitude higher total resolution.

Addressing these challenges, our work explores scene reconstruction using active gated imaging. Gated imaging functions by integrating the transient response from a scene that has been flash-illuminated by a synchronized light source \cite{busck2004gated}. This imaging technique is robust to adverse weather conditions -- temporal gating allows us to filter out backscatter -- and provides signal in poorly-lit scenes \cite{bijelic2018benchmarking}. Existing work has exploited this sensing modality to achieve state-of-the-art depth estimation \cite{gated2gated,gatedstereo} as well as object detection \cite{julca2021gated3d}.
In our approach, we train a neural field-based representation of the scene, concurrently learning its geometry, illumination, and material properties. This is accomplished by integrating the gated imaging formation model with a neural rendering framework, that jointly learns the associated gating parameters along with the scene reconstruction. By leveraging the implicit depth cues present in gated video captures, we are able to reconstruct a detailed 3D geometric model of the scene, as shown in \cref{fig:teaser}.
Compared to LiDAR-based approaches, our method offers distinct advantages, as LiDAR systems are inherently constrained by their resolution, necessitating additional time-multiplexed scene captures to aggregate points. This results in an extended acquisition process for LiDAR-based methods or, conversely, compromises the geometric detail of the final estimate. Specifically, for a fixed time acquisition budget the scene reconstruction from a LiDAR sensor is less supervised, although offering highly accurate depth information, the sensor yields data at a volume one order of magnitude less than that of a camera stereo pair. This disparity in data quantity means that while the LiDAR provides precise depth points, it is unable to provide dense and fine detailed predictions.

To assess our method, we captured a dataset of varied scenes at day and night conditions, using a vehicle test setup comprising of LiDAR, RGB and Gated sensors. We compare our method with feed-forward and 3D reconstruction methods, and assess its superiority in novel depth synthesis beating the next best method by 21.87$\%$ MAE, 3D reconstruction improving on the baseline by 11$\%$ IoU, and performs novel view synthesis with a PSNR of 32.28 dB.

\vspace{5pt}
\noindent
In our work, we make the following contributions:

\begin{itemize}
  \item We propose a novel neural rendering method and scene representation that is capable of reconstructing scene geometry and radiance from active gated camera videos.

  \item By modeling the gated image formation process and integrating into differentiable volume rendering, our approach is able to reconstruct and decompose both passive and active light transport components conditioned on the scene parameters in a physically accurate way.
 
  \item  We validate our method on large outdoor scenes, captured across different scenarios in both day and night, achieving a reduction of MAE error in depth precision of $59.8\%$ to the next best RGB+LiDAR method and $31.7\%$ to the next best methods using gated images.
\end{itemize}
\section{Related Work}
\label{sec:formatting}

\PAR{Monocular and Stereo Depth Estimation}
Depth estimation tasks, from a single image \cite{MovingPeopleMovingCameras, godard2017unsupervised,li2022depthformer,guizilini20203d}, from stereo image pairs \cite{chang2018pyramid,TransformerStereo,badki2020Bi3D,yang2019hsm}, or single/stereo images with a LiDAR scan \cite{tang2019sparse2dense, wong2021unsupervised, park2020nonRGBLidar, hu2020PENetRGBLidar, guidenetRGBLidar,VolPropagationNetStereoLidar, SLFNetStereoLidar} have been at the center of a large body of work. 
Single CMOS sensor-based depth estimation from RGB color images is inherently limited by scale ambiguity. Additional measurements from LiDAR \cite{bartoccioni2023lidartouch} or ego-vehicle speed \cite{guizilini20203d} can resolve this ambiguity at the cost of an additional sensor. Similarly, stereo methods rely on an additional camera sensor to resolve ambiguity by triangulating between two camera views \cite{chang2018pyramid}. Training approaches for learned depth estimation methods using intensity images cover both unsupervised methods \cite{Zhou2017, godard2019digging,guizilini20203d, Garg2016, godard2017unsupervised}, which harnesses multi-view geometry consistency, and supervised techniques \cite{li2022depthformer, eigen2014depth,chang2018pyramid,jaritz2018sparse,ma2018sparse,MovingPeopleMovingCameras,Mayer2016,Kendall2017} relying primarily on multi-view datasets \cite{MovingPeopleMovingCameras,Mayer2016,Kendall2017} or time-of-flight captures \cite{eigen2014depth, chang2018pyramid,jaritz2018sparse,ma2018sparse}. In particular, LiDAR measurements have been proven as a ground-truth signal for depth supervision \cite{agarwal2023attention,li2022depthformer, eigen2014depth,chang2018pyramid,jaritz2018sparse,ma2018sparse}. Several methods~\cite{geiger2012we,Uhrig2017THREEDV} have mitigated the sparsity and range limitations of scanning LiDAR by accumulating scans. However, adverse weather \cite{Bijelic_2020_STF} can make LiDAR ground truth unreliable, and methods that rely on consistency between camera and LiDAR \cite{tang2019sparse2dense, wong2021unsupervised, park2020nonRGBLidar, hu2020PENetRGBLidar, guidenetRGBLidar, VolPropagationNetStereoLidar, SLFNetStereoLidar} suffer from degradations, including scan pattern artifacts and temporal distortions.

\PAR{Time-gated Depth Sensing}
Time-of-Flight (ToF) sensors determine depth by emitting light into a scene and calculating the distance based on the round trip time of the light. Successful sensing methods can be categorized into three classes: correlation ToF cameras \cite{hansard2012time, kolb2010time, lange00tof}, pulsed ToF sensors \cite{schwarz2010lidar}, and gated imaging \cite{heckman1967,grauer2014active}.
Correlation ToF sensors\cite{hansard2012time, kolb2010time,
lange00tof} estimate depth by continuously flood-illuminating the scene and assessing the phase shift between the emitted and received light.
While these sensors can provide high-resolution depth information, their use is primarily confined to indoor settings due to their susceptibility to external light interference.
Pulsed ToF sensors \cite{schwarz2010lidar} function by emitting light pulses toward specific scene points and measure the total travel time to estimate depth. Emitting collimated light, this approach is robust against ambient illumination and allows for outdoor depth measurements. However, its spatial resolution is fundamentally limited owing to its scanning illumination technique, and its efficacy can be compromised in adverse weather due to backscatter \cite{BenchmarkLidar,LIBRE,Jokela}.
In contrast, gated cameras ~\cite{heckman1967,grauer2014active,Bijelic2018} capture light from a scene over brief intervals, essentially constraining the observable depth to specific range segments. The inherent gating mechanism of these cameras offers resistance to backscattering, and they allow for the recovery of detailed depth maps when using a large number of short gates \cite{Busck2004, Busck2005, Andersson2006} 
Subsequent works have improved gated depth estimation with few gates by adopting Bayesian approaches \cite{adam2017bayesian,schober2017dynamic} or deep neural networks \cite{gated2depth2019, gated2gated, gatedstereo}, and they achieve accurate gated depth estimation for dynamic outdoor scenes, even under challenging conditions. Recently, Gated Stereo \cite{gatedstereo} reached state-of-the-art results using a stereo-gated setup and self-supervision~\cite{gated2gated}.

\PAR{Neural Scene Reconstruction}
Recent research has amalgamated sets of single-sensor measurements to recover comprehensive scene representations. This synthesis has led to advancements in both novel-view generation \cite{mildenhall2020nerf,barron2021mipnerf,chen2022tensorf,muller2022instantNGP} and depth estimation \cite{tosi2023nerf4depth} , with neural radiance field methods \cite{mildenhall2020nerf,barron2021mipnerf,chen2022tensorf,muller2022instantNGP} emerging as a pivotal approach for representing scenes as continuous volumetric fields of radiance.
These methods combine this representation with volumetric rendering as a forward model in a test-time optimization approach. Specific representations that these methods explore include coordinate-based networks \cite{mildenhall2020nerf,barron2021mipnerf,barron2022mipnerf360,zhang2021nerfactor}, 3D voxel-grid representation \cite{fridovich2022plenoxels,yu2021plenoctrees,chen2022tensorf}, or hybrid approaches \cite{muller2022instantNGP,barron2023zipnerf,tancik2023nerfstudio}. 
Subsequent works have extended this representation to large outdoor scenes \cite{zhang2020nerf++,barron2022mipnerf360}, and increased the efficiency at training and test time \cite{yu2021plenoctrees,barron2023zipnerf,chen2022tensorf,muller2022instantNGP}. Other works departed from radiance-based representation and explicitly learned the scene illumination, geometry, and material proprieties \cite{zhang2021nerfactor, boss2022samurai, rudnev2022nerfOSR}. A particular challenge within the field is reconstructing large urban terrains based on imagery captured from vehicles \cite{tancik2022blocknerf,turki2023suds,rematas2022urf,guo2023streetsurf,wang2023fegr,kundu2022panoptic,yang2023unisim,liu2023waabiICCV23,ost2022pointlightfields}, given that a significant portion of the scene is seen from only a narrow range of viewpoints. This problem is tackled by additional supervision cues from sparse LiDAR \cite{ost2022pointlightfields, turki2023suds, rematas2022urf,guo2023streetsurf}, pre-estimated depths \cite{deng2022depthNeRF,roessle2022densedepthNeRF,guo2023streetsurf}, optical flow \cite{turki2023suds,meuleman2023localrf} and semantic segmentation \cite{kundu2022panoptic, wang2023fegr,turki2023suds}.   
Departing from RGB-based approaches, recent works have investigated neural reconstruction methods using ToF sensors \cite{malik2023transientnerf,huang2023nfl, tao2023lidarnerf}. Existing methods \cite{huang2023nfl, tao2023lidarnerf,zhang2023nerflidar} learn a neural field from posed LiDAR scans, allowing for the synthesis of realistic LiDAR scans from novel views. Recently, Malik et al.~\cite{malik2023transientnerf} represent the time-resolved photon count acquired by a single-photon LiDAR system with a neural reconstruction method.

\section{Gated Imaging Model}
\label{sec:gated_imaging_model}

\begin{figure}[!t]
\vspace{-1mm}
    \centering
    \includegraphics[width=0.45\textwidth]{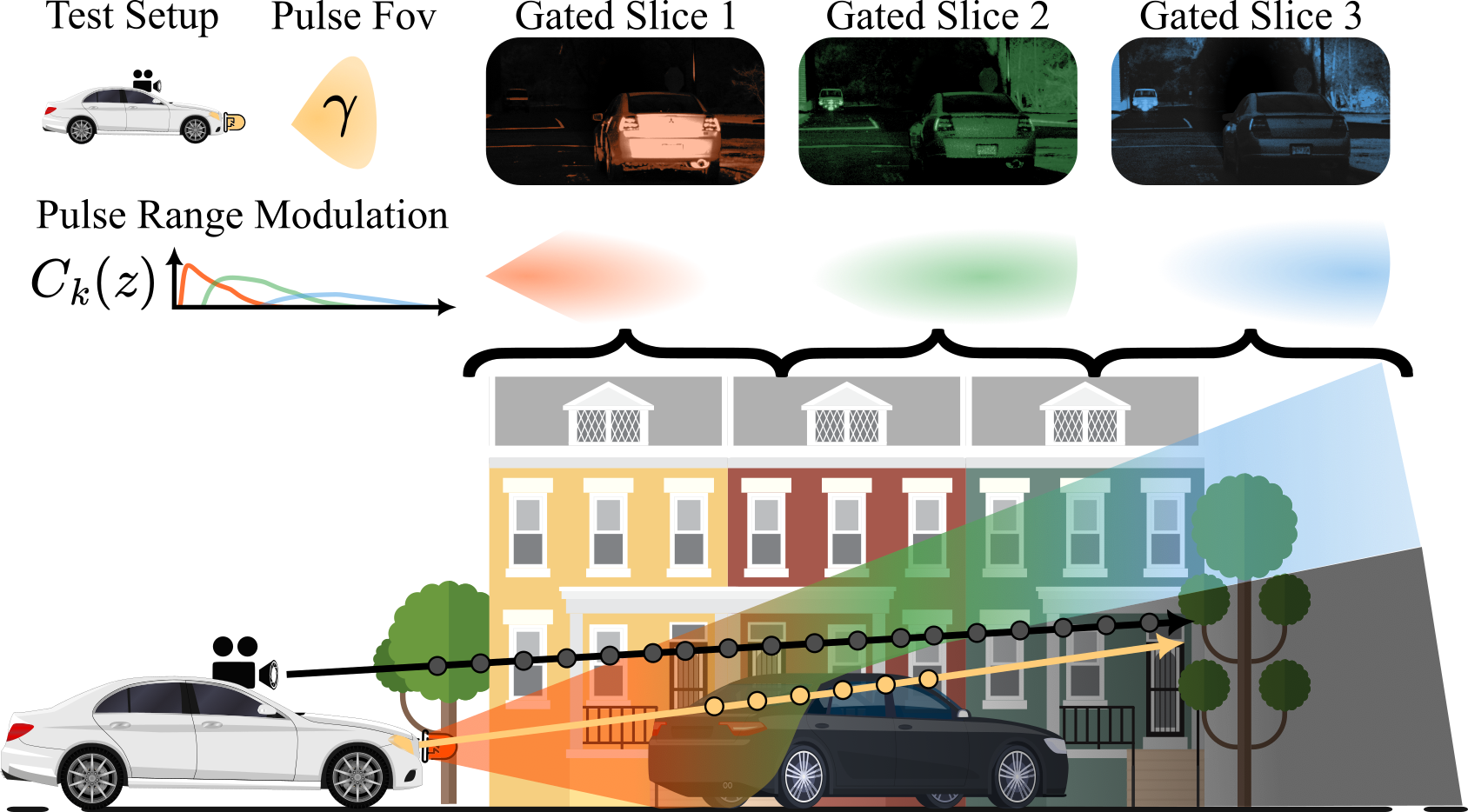}
    \caption{ Gated Image Formation and Bi-Directional Sampling. Top-row: Our test vehicle is equipped with a synchronized stereo camera setup and illuminator that flood-lits the scene with a light pulse and FoV $\gamma$. Using different gating profiles $C(z)$, we capture three slices with intensity visualised here in red, green and blue. Illustrated in the middle row, the gating profiles describe pixel intensity for a point at sensor distance $z$. The first slice (red) accounts for close ranges, the green for mid-ranges, and the blue for far ranges. Bottom-row: we show the ray sampling employed in our method, based on a bidirectional sampling strategy. We cast the rays from the illuminator view to explore the occluded areas, while the rays casted from the camera integrate the reflected scene response. The shadowed areas are marked in gray. 
    }
    \label{fig:arch_2f}
    \vspace{-3mm}
\end{figure}

In this section, we provide a brief background on gated imaging as presented in~\cite{gated2depth2019} and introduce the proposed gated image formation model. Unlike prior work, this model incorporates shadow effects and features self-calibrated parameter learning.

A gated imaging system, illustrated in Fig.~\ref{fig:arch_2f}, utilizes a pulsed flood-light illumination source $p$ with a synchronized imager that operates with a nanosecond (ns) gated exposure \( g \) that is delayed by \( \xi \) compared to the pulse. This allows us to capture only photons with round-trip times inside the gates, hence specific distance segments in the scene. We formalize this using so-called range intensity profiles \( C_k(2z_c) \) given distance \( z_c \) from the camera, time \( t \), and a parameter set \( k \)~\cite{gated2depth2019}, that is
\begin{equation}
    C_k(2z_c)=\int\limits_{-\infty}^{\infty} g_k(t-\xi)p_k\left(t\,-\,\cfrac{2z_c}{c}\right)\beta(2z_c)dt,
\label{eq:integral_gated_basic}
\end{equation}
where $c$ is the speed of light and ~$\beta(\cdot)$ models the distance-dependent decay of the reflected light pulse. The resulting gated pixel value is
\begin{equation}
    I_{k}(z_c) = \alpha \iota C_k(2z_c) + \Lambda + \mathcal{D}_k,
    \label{eq:toteq_gated_basic}
\end{equation}
where \( \Lambda \) represents the passive ambient contribution, ~$\alpha$ is the scene reflection, ~$\iota$ the laser illumination, and \( \mathcal{D}_k \) is an additive noise term.

This model assumes camera position \( \mathbf{o}_c \) and the illuminator position \( \mathbf{o}_i \) are collocated. To allow for non-collocated positions, we express the travel time as \( z = z_c + z_i \), where \( z_i \) denotes the distance between the illuminator and the point on the surface impacted by the light beam, represented as \( z_i = |\mathbf{x} - \mathbf{o}_i|_2 \). Additionally, there may be areas visible to the camera that remain dark due to potential occlusions. Modeling shadow effects and attenuation due to incident angle $\omega$ results in the following image formation
\begin{equation}
\small I_{k}(z) = \alpha \iota \psi |\mathbf{n} \cdot \omega | C_k(z)+ \Lambda + \mathcal{D}_k.
\label{eq:gated_img}
\end{equation}
Here, $\psi \in [0,1]$ serves as a shadow indicator for the pixel, and $\omega$ is the direction of the incident light at that point.

We extend this model to fit the range intensity profiles during optimization, thereby eliminating the need for their direct measurement. This approach overcomes potential calibration inaccuracies encountered in previous approaches \cite{gated2gated,gatedstereo}. We model both the laser pulse $p_k$ and the gate $g_k$ as rectangular functions with durations $t_{l,k}$ and $t_{g,k}$, respectively. This simplification permits the analytical computation of the integral in \cref{eq:integral_gated_basic}, that is
\begin{equation}
    \tilde{C}_k \! = \!  
    \begin{cases}
      \frac{2z}{c}-\xi_k + t_{l,k} & \text{if } \xi_k - t_{l,k} < \frac{2z}{c}< \xi_k \\
      t_{l,k} & \text{if } \xi_k < \frac{2z}{c} < \xi_k + t_{g,k} - t_{l,k} \\
      \! -\frac{2z}{c} +\!  \xi_k \!  + t_{g,k} & \text{if } \xi_k + l_{g,k} - t_{l,k} < \! \frac{2z}{c} \! < \xi_k \! + t_{g,k} \\
      0 & \text{otherwise}
    \end{cases}\
\label{eq:gating_equation_simpl}    
\end{equation}

\begin{figure*}[!t]
\vspace{-3mm}
    \centering
    \includegraphics[width=0.85\textwidth]{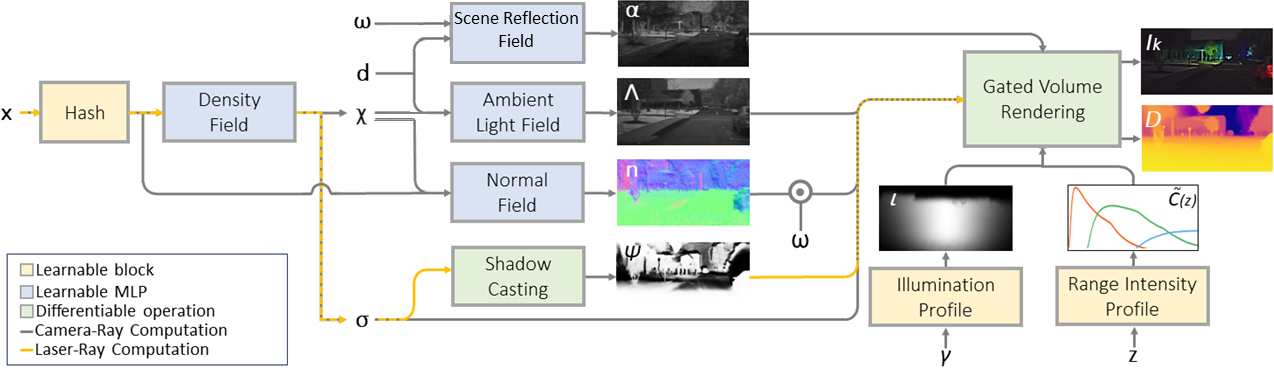}\vspace{-8pt}
    \caption{ Neural Gated Fields. For any point in space $\mathbf{x}$, we learn its volumetric density $\sigma$, normal $\mathbf{n}$, reflectance $\alpha$ and ambient lighting $\Lambda$ through four neural fields, conditioned on direction $\mathbf{d}$, incident laser light direction $\mathbf{\omega}$ and spatial embedding $\chi$. The illuminator light $\iota$ is represented by a physics-based model dependent on the displacement angle $\gamma$, while the gating imaging process is described by the range intensity profiles $\tilde{C}(z)$ using as input the camera-point-laser distance $z$, as explained in \cref{sec:gated_imaging_model}.
With this information, we reconstruct a gated image $I_k$ through the gated volume rendering formulation introduced in \cref{gated_field_learning}. As this process is fully differentiable, we simultaneously fit neural fields and physical parameters through image reconstruction together with other regularization losses discussed in \cref{training_supervision}. 
    }
    
    \label{fig:arch}
    \vspace{-3mm}
\end{figure*}

\vspace{-3mm}
\section{Gated Field}
\label{neural_gated_field}
We reconstruct a scene by fitting a neural field representation to gated videos. We collect videos of three active gated slices $I_{k\in\{1,2,3\}}$ with different gating parameters and one passive slice $I_P$. We model active illumination by jointly estimating light and material properties, and separately represent the ambient light as a radiance field. The proposed reconstruction method relies on both photometric reconstruction cues and scene priors and is described in the following.

\subsection{Neural Gated Fields}
We describe the scene proprieties using two neural fields $f_{Gp}$ and $f_{G\alpha}$, respectively representing the ambient light scattered in the scene and the reflection of the scene surfaces, conditioned on a spatial embedding $\chi$.
Moreover, the laser illumination contribution is represented by a physics-based model, while shadow effects are simulated through ray-tracing using the volumetric density field $f_{Gd}$.
\paragraph{Neural Ambient and Reflection Field}
\label{par:neural_ambient_and_reflectance_field}
We represent a scene as a neural field $f_G \! : \! \{ \mathbf{x},\mathbf{d}, \mathbf{\omega} \} \! \! \rightarrow \! \!\{ \sigma,\alpha, \Lambda, \mathbf{n} \}$ mapping each point in space $\mathbf{x}$ viewed from a direction $\mathbf{d}$ and laser direction $\omega$ to its volumetric density $\sigma$, normal vector $\mathbf{n}$,  scene reflection $\alpha$ and the passive component $\Lambda$, that is

\begin{equation*}
\begin{aligned}
\! f_{Gp} &: \{ \mathbf{d}, \chi \} \relbar\joinrel\relbar\joinrel\rightarrow \{ \Lambda \} & \! \text{Ambient Component} \\
\! f_{Gn} &: \{ \mathbf{x}, \chi \} \relbar\joinrel\relbar\joinrel\rightarrow \{ \mathbf{n} \} & \! \text{Surface Normal} \\
\! f_{G\alpha} &: \{ \mathbf{d}, \mathbf{\omega}, \chi \} \rightarrow \, \{ \alpha \} & \! \text{Surface Reflection} \\
\text{with}\quad \! \! \! \! f_{Gd} &: \{ \mathbf{x} \} \: \DOTSB\relbar\joinrel\relbar\joinrel\rightarrow  \{ \sigma, \chi \} & \! \text{Vol. Density and Embedding}
\end{aligned}
\end{equation*}

We condition here normal, ambient and reflectance on a volumetric embedding via the field $f_{Gd}: \{ \mathbf{x} \} \rightarrow \{ \sigma, \chi \} $ estimating the density $\sigma$ and embedding $\chi$. This embedding is being shared by the network branches to estimate the normal with $f_{Gn}: \{ \mathbf{x}, \chi \} \rightarrow \{ \mathbf{n} \}$, scene reflection $f_{G\alpha} : \{ \mathbf{d}, \mathbf{\omega}, \chi \} \rightarrow \{ \alpha \}$, and ambient light component with $f_{Gp} : \{ \mathbf{d}, \chi \} \rightarrow \{ \Lambda \}$. An overview of the overall Neural Gated Fields is shown in \cref{fig:arch}.
We also use a proposal sampler $f_P$ as in \cite{barron2022mipnerf360} for efficiency. Both $f_G$ and $f_P$ are MLPs (of different size) with multi-resolution hash encoding \cite{muller2022instantNGP}.

\paragraph{Shadow and Illumination Model}
As the light pulse emitted by the illuminator is a diverging light beam, we model it as cone of light with irradiance maximum at the cross-section center and exponentially decreasing as it diverges from the center by the angles $\gamma$. As such, we express the illumination intensity as a 2D higher-order Gaussian $\mathcal{G}$ with mean $\mathbf{\Xi}$, standard deviation $\mathbf{\Omega}$ and power $\mathbf{\Theta}$
\begin{equation}
\iota = \eta \mathcal{G}_i (\mathbf{\gamma};\mathbf{\Xi},\mathbf{\Omega}, \mathbf{\Theta}),
\label{eq:iota_equation}
\end{equation}
where $\eta$ is a scaling parameter.

Instead of predicting the shadow indicator $\psi (\mathbf{x})$, we can directly estimate it using the density field, by computing the accumulated transmittance along the ray from the pixel to the point $\mathbf{r}_{ill} (l)= \mathbf{o}_i + \mathbf{\omega} l$
\begin{equation}
\psi(\mathbf{x}) = \exp \left( - \int_0^{l_{\mathbf{x}}} \sigma(\mathbf{r}_{ill}(l)) \textrm{d}l \right)
\label{eq:psi_equation}
\end{equation}
The illuminator origin and direction $\mathbf{o}_i$, $\omega$ are obtained from the camera as $[\mathbf{o}_i$, $\omega] =\mathbf{R}[\mathbf{o}_c,\mathbf{d}_c]+\mathbf{T}$. During training, we jointly fine-tune the translation and rotation matrices $\mathbf{T}$, $\mathbf{R}$, as well as $\mathbf{o}_c$. We also treat illuminator profile proprieties as learnable parameters, namely $\eta$, $\mathbf{\Xi}$, $\mathbf{\Omega}$, $\mathbf{\Theta}$ and the gating parameters, i.e. number of accumulated laser pulses $m_k$ before read-out, laser pulse duration $t_{l,k}$, camera exposure $t_{g,k}$, and delay $\xi_k$ between laser pulse emission and gated exposure for all three slices $k \in \{0, 1, 2\}$. In addition, we optimize a general distance offset $d_0$ for the range intensity profiles to compensate for internal signal processing delays.

\subsection{Gated Field Learning}
\label{gated_field_learning}
We learn to acquire a gated capture with camera origin $\mathbf{o}_c$ and direction $\mathbf{d}$ by casting a ray $\mathbf{r}(l)=\mathbf{o_c}+l\mathbf{d}$ for each pixel into the scene and computing the intensity $\tilde{I}_k(\mathbf{r})$ through volume rendering. Using the gated imaging formation model from \cref{sec:gated_imaging_model}, we define volume rendering as
\begin{equation}
\begin{split}  
\tilde{I}_{k}(\mathbf{r}) = \int_0^{\infty} T(l) \sigma (\mathbf{x})
\int_{-\infty}^{\infty}  g_k(t-\xi_k) \beta(l)\\
\cdot \left( \alpha \iota \psi |\mathbf{n} \cdot \omega | p_k\left(t-\frac{l+l_i}{c}\right) + \kappa \right) \textrm{d}t \, \textrm{d}l + \mathcal{D}_k,
\end{split}
\label{eq:gating_integral_neural}
\end{equation}
where $\kappa$ is the ambient level, $T(l) = \exp(-\int_0^l \sigma(u) du )$ is the accumulated transmittance along the ray and $l_i$ is the distance of $\mathbf{x}(l)$ from the illuminator origin $\mathbf{o}_i$. 
As illustrated in \cref{fig:arch_2f}, in our volume rendering formulation the pixel intensity contribution of a point along the ray depends not only on its accumulated transmittance and volumetric density, but also on its distance from the illuminator and camera origins through $C(z)$, as well as on its relative position to the illuminator source via $\iota$ and $\psi$.\\
We simplify the time-dependent integral using \cref{eq:gating_equation_simpl} as
\vspace*{-6pt}
\begin{equation}
\begin{aligned}  
\tilde{I}_{k}(\mathbf{r}) = \int_0^{\infty} T(l) \sigma (\mathbf{x}) \big(
\alpha(\mathbf{x},\mathbf{d}, \mathbf{\omega})  \tilde{C}_k(l+l_i) \psi(\mathbf{x}) \\ \cdot |\mathbf{n} \cdot \omega | \iota( \gamma )  
 +\Lambda(\mathbf{x},\mathbf{d}) \big) \textrm{d}l + \mathcal{D}_k.
\label{eq:gating_integral_neural}
\end{aligned}
\end{equation}
We numerically estimate this spatial integral by numerical quadrature~\cite{mildenhall2020nerf,verbin2022refnerf}, approximating it with a set of points $\mathbf{X}_{ray}$. Specifically, for each point $\mathbf{x}_j \in \mathbf{X}_{ray}$, we query the neural field $f_G$ to infer the normal vector $\mathbf{n}_j$, scene reflection $\alpha_j$, volumetric density $\sigma_j$ and ambient light component $\Lambda_j$. The laser illumination intensity $\iota_j$ is instead computed following the physics-based model defined in \cref{eq:iota_equation}. The gated intensity $\tilde{I}_{k}$ is then expressed as
\vspace*{-8pt}
\begin{equation}
\tilde{I}_{k}(\mathbf{r}) = \sum_{j=0}^{N} w_j \Big( \underbrace{\alpha_j \Tilde{C}_{j} \psi_j |\mathbf{n}_j \cdot \omega_j | \iota_j}_{\substack{\text{Active} \\ \text{Component}}} +\underbrace{\Lambda_j}_{\substack{\text{Passive} \\ \text{Component}}} \Big)  +\mathcal{D}_k,
\label{eq:gating_integral_neural}
\end{equation}
\vspace*{-14pt}
\begin{equation}
w_j = \exp(-\sum_{k=1}^{j-1}\sigma_k \delta_k) (1-\exp(-\sigma_j\delta_j) ).
\label{eq:weights_appr}
\end{equation}
The shadow indicator $\psi_j$ from \cref{eq:psi_equation} is similarly approximated by sampling on $\mathbf{r}_{ill} = \mathbf{o}_i + \omega_j l$ a set of points $\mathbf{X}_{ill}$ bounded between $\mathbf{o}_i$ and $\mathbf{x}_j$
\vspace*{-8pt}
\begin{equation}
\psi_j = \exp(-\sum_{k}^{\mathbf{X}_{ill}}\sigma_k \delta_k).
\end{equation}
For the passive slice, the active component is null, further simplifying to
\vspace*{-8pt}
\begin{equation}
\tilde{I}_{P}(\mathbf{r}) = \sum_{j=0}^{N} w_j \Lambda_j +\mathcal{D}_P.
\label{eq:gating_integral_neural_passive}
\end{equation}
Both $\mathbf{X}_{ray}$ and $\mathbf{X}_{ill}$ are sampled using a proposal network \cite{barron2022mipnerf360} $f_P$ that, analogously to $f_{Gd}$, predicts point-wise densities converted with \cref{eq:weights_appr} to proposal weights $\hat{w}$ for sampling with piece-wise-constant probabilities.     

\subsection{Training Supervision}
\label{training_supervision}
\vspace{-6pt}
We supervise the predicted passive and active gated frames applying a photometric loss $\mathcal{L}_c$, regularize the volumetric density with a depth loss $\mathcal{L}_d$ and by supervising the shadow estimate with $\mathcal{L}_s$. We regularize normal and reflectance estimates through $\mathcal{L}_{nc}$ and $\mathcal{L}_\alpha$, respectively. \\
\vspace{-12pt}
\PAR{Photometric Loss}
We supervise with ground truth captures for active and passive gated slice reconstruction as
\vspace{-8pt}
\begin{equation}
\mathcal{L}_{c}  = \sum_{k,r}\Vert \tilde{I}_k(\mathbf{r}) - I_k(\mathbf{r}) \Vert_2 + \sum_r \Vert \tilde{I}_P(\mathbf{r}) - I_P(\mathbf{r}) \Vert_2 
\label{eq:gating_integral_neural}
\end{equation}
\vspace{-8pt}
\PAR{Volume Density Regularization}
As additional training supervision, we use the depth estimate $\hat{D}(\mathbf{r})$ of a pretrained stereo depth estimation algorithm \cite{gatedstereo} as pseudo ground-truth to regularize the ray termination distribution \cite{deng2022depthNeRF}
\vspace{-8pt}
\begin{equation}
\mathcal{L}_{d} = \sum_{\mathbf{r}} \sum_j \text{log}w_j \exp \left( - \frac{(l_j - \hat{D}(\mathbf{r}))^2}{2s^2} \right) \delta_i
\label{eq:depth_loss}
\end{equation}

We regularize the density field by partially supervising the shadow indicator $\psi$. Each pixel whose active intensity $I_{kA} = I_k- I_P$ in any of the three gated slices is above a certain threshold $\epsilon_i$ is considered as visible from the illuminator. We hence supervise the expected shadow value for such rays $\mathbf{r}_v \in \{\mathbf{r} | \forall k \in \{1,2,3\}: I_{kA}(\mathbf{r}) > \epsilon_i \}$ as
\vspace{-8pt}
\begin{equation}
\mathcal{L}_{s} = \sum_{\mathbf{r}_v} \Vert 1- \int T(l) \sigma(\mathbf{x}) \psi(\mathbf{x}) dl \Vert _2 
\label{eq:shadow_loss}
\end{equation}

\vspace{-8pt}
\PAR{Normals Consistency}
Following \cite{verbin2022refnerf}, for each sampled point $\mathbf{x}$ we enforce a consistency between the predicted normal $\mathbf{n}$ and the density gradient $\hat{\mathbf{n}}(\mathbf{x}) = -\frac{\nabla \mathbf{x}}{||\nabla \mathbf{x}||}$, and we penalize normals which are back-facing the camera as
\vspace{-7pt}
\begin{equation}
\! \! \! \mathcal{L}_{nc}\!  =\!  \sum_{\mathbf{x}}\!  w(\mathbf{x}) \! \left( \Vert \mathbf{n}(\mathbf{x}) \! - \! \hat{\mathbf{n}}(\mathbf{x}) \Vert_2 \! + \! \max(0,\hat{\mathbf{n}}(\mathbf{x})\! \cdot \! \mathbf{d})^2 \right)
\label{eq:normal_consistency}
\end{equation}

\vspace{-8pt}
\PAR{Reflectance Regularization }
We enforce the predicted scene reflection $\alpha$ to be spatially consistent within $\mathbf{\epsilon_{\mathbf{x}}}$, i.e.
\vspace{-7pt}
\begin{equation}
\mathcal{L}_{\alpha} = \sum_{\mathbf{x}} w(\mathbf{x}) \left( \Vert \alpha(\mathbf{x},\mathbf{d}) - \alpha(\mathbf{x}+\mathbf{\epsilon_x},\mathbf{d}+\mathbf{\epsilon_d}) \Vert_2  \right) .
\label{eq:reflectance_reg_loss}
\end{equation}
Here the $\mathbf{\omega}$ dependency of $\alpha$ is omitted here for brevity. We also include an angular noise $\mathbf{\epsilon_d}$ that we set high at the beginning of the training and then decrease it exponentially. By forcing the reflectance to behave as fully diffuse at the beginning of the training, we disincentive it to bake in the effects from lighting or shadowing, hence improving the disjoint learning of the scene components.
\PAR{Total Training Loss}
Combining the different losses we obtain the following loss formulation:
\begin{equation}
\mathcal{L} = \lambda_{1} \mathcal{L}_{c} +\lambda_2 \mathcal{L}_d +\lambda_3 \mathcal{L}_s +\lambda_4 \mathcal{L}_{nc} + \lambda_5 \mathcal{L}_{\alpha},
\label{eq:tot_loss}
\end{equation}
see $\lambda_{1,...,5}$ hyperparameters in the Supplementary Material.

\vspace{-2mm}
\section{Implementation Details}
\vspace{-2mm}
We train for 35,000 steps and a batch size of 4096 rays. As optimizer we use ADAMW \cite{ADAMW} with $\beta_1\,=\,0.9$, $\beta_2\,=\,0.999$, learning rate $10^{-2}$ for $f_P$ and $f_G$, $10^{-4}$ for the camera poses optimization, $10^{-4}$ for the laser profile and gated parameters. We train on two NVIDIA V100 GPUs, for approximately 3 hours. The proposal network $f_P$ is comprised of two MLPs and trained following \cite{barron2022mipnerf360}. Additional architecture details, training procedures, and hyper-parameters are found in the Supplementary Material. 
\vspace{-2mm}
\section{Dataset}
\label{dataset}
\begin{figure}[!t]
    \vspace{-3mm}
    \centering
\noindent\includegraphics[width=0.49\textwidth]{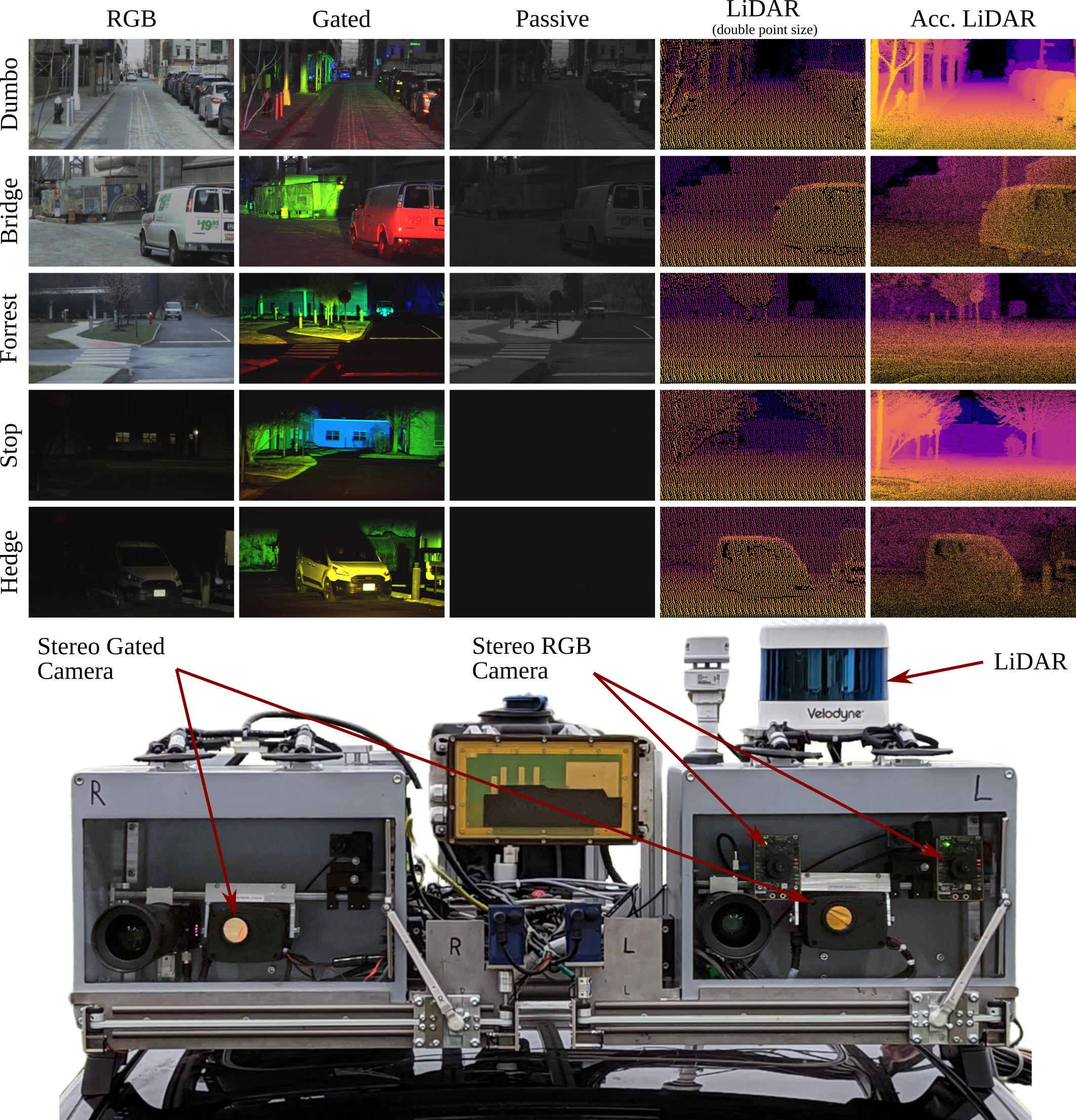} 
    \vspace{-4mm}
    \caption{
    Top: Example captures from our collected dataset across different urban and suburban areas in North America. From left to right: RGB image, active gated slices (with red for slice 1, green for slice 2 and blue for slice 3), passive slice, projected LiDAR scan, accumulated LiDAR. Bottom: 
    Sensors setup with LiDAR, stereo Gated camera, stereo RGB camera, IMU and GNSS.}
    \label{fig:dataset}
    \vspace{-4mm}
\end{figure}

\vspace{-2mm}
To conduct this work, we have collected a diverse set of 10 static sequences, recorded in both day and night conditions across North America. To this goal, we equipped a test vehicle with a NIR gated stereo camera setup (BrightWay Vision), an automotive RGB stereo camera (OnSemi AR0230), a LiDAR sensor (Velodyne VLS128) and a GNSS with IMU (Xsens MTi-7), as shown in \cref{fig:dataset}. 
Each gated camera has a resolution of 1280x720 pixels, \unit[10]{bit} depth and runs at \unit[120]{Hz}, split up to collect the three active and one passive slice. The illuminator source consists of two vertical-cavity surface-emitting laser (VCSEL) modules, which illuminate the scene with a laser pulse with duration of \unit[240-370]{ns} and a wavelength of \unit[808]{nm}. 
The RGB cameras provide \unit[12]{bit} HDR images with resolution of 1920x1080 pixels and \unit[30]{Hz} frame-rate. The LiDAR has a vertical resolution of 128 lines and \unit[10]{Hz} framerate, while the GNSS sensor runs at \unit[4]{Hz}. 
Example captures from the dataset are being visualized in \cref{fig:dataset}. In total, we collect 2650 samples, captured in both day (1223 samples) and night (1427 samples). We divide it in training, validation and test splits with a 50-25-25 split. 

As ground-truth, we construct a large-scale ground-truth pointcloud by aggregating LiDAR scans with LIO-SAM \cite{shan2020liosam} and removing noisy points. Additional details on the accumulation are provided in the Supplementary Material.

\vspace{-2mm}
\section{Assessment}
\vspace{-2mm}
In this section, we validate the proposed method quantitatively and qualitatively. Specifically, we investigate scene reconstruction at both day and night, using novel depth and view synthesis for 2D evaluation, and surface reconstruction for the 3D evaluation. To this end, we compare our approach to state-of-the-art feed-forward depth estimation algorithms and neural scene reconstruction methods. We also conduct ablation experiments to validate our design choices. 

\begin{table}[!t]
    \vspace{-0.8eM}
    \footnotesize
    \setlength{\tabcolsep}{4pt} 
    \setlength\extrarowheight{2pt}
    \centering
    \resizebox{.99\linewidth}{!}{
    \begin{tabular}{@{}c|lccccccc@{}}
            \toprule
             & \multirow{2}{*}{\textbf{\textsc{Method}}} & \textbf{Modality}  & \textbf{RMSE}     & \textbf{ARD}   & \textbf{MAE}  & $\boldsymbol{\delta_1}$ & $\boldsymbol{\delta_2}$ & $\boldsymbol{\delta_3}$  \\ 
			&&   & $\left[ m \right]$  &  & $\left[ m \right]$ & $\left[ \% \right]$ & $\left[ \% \right]$ & $\left[ \% \right]$\\
			\midrule
			\multicolumn{9}{c}{\textbf{Test Data -- Night (Evaluated on Accumulated LiDAR Ground-Truth Points)}} \\
			\midrule
			\multirow{16}{*}{\rotatebox[origin=l]{90}{\parbox[c]{6cm}{\centering \textbf{\textsc{2D Depth Comparison}}}}} 
			 & \textsc{Gated2Gated}  \cite{gated2gated} & Gated & 13.33  &  0.28  &  8.57  &  61.78  &  90.30  &  94.90  \\ 
 & \textsc{GatedStereo} \cite{gatedstereo} & Stereo-Gated & \underline{10.10 } &  0.20  &  5.97  &  82.86  &  93.35  &  96.52  \\ 
 & \textsc{SimIPU} \cite{li2022simipu} & RGB & 19.33  &  0.44  &  14.21  &  40.67  &  77.99  &  89.80  \\ 
 & \textsc{AdaBins} \cite{bhat2021adabins} & RGB & 21.14  &  0.38  &  14.28  &  51.72  &  80.00  &  90.64  \\ 
 & \textsc{DPT} \cite{ranftl2021vision} & RGB & 14.17  &  0.28  &  9.90  &  62.82  &  88.32  &  94.42  \\ 
 & \textsc{DepthFormer} \cite{li2022depthformer} & RGB & 14.32  &  0.29  &  10.08  &  61.19  &  87.72  &  94.57  \\ 
 & \textsc{CREStereo} \cite{li2022practical} & Stereo-RGB & 14.22  & \underline{ 0.13 } &  7.27  &  82.21  &  90.10  &  94.32  \\ 
 & \textsc{NLSPN} \cite{park2020nonRGBLidar} & RGB+LiDAR & 11.12  &  0.18  &  6.48  &  77.15  &  90.81  &  96.00  \\ 
 & \textsc{MipNeRF360} \cite{barron2022mipnerf360} & RGB & 23.80  &  0.51  &  16.43  &  41.60  &  61.75  &  76.16  \\ 
 & \textsc{K-Planes} \cite{fridovich2023kplanes} & RGB & 19.70  &  0.41  &  13.66  &  44.39  &  66.20  &  81.46  \\ 
 & \textsc{RawNeRF} \cite{mildenhall2022nerfdark} & RGB & 27.46  &  0.64  &  19.75  &  34.32  &  54.33  &  69.41  \\ 
 & \textsc{Depth-NeRF} \cite{deng2022depthNeRF} & RGB & 15.23  &  0.26  &  10.04  &  61.67  &  86.23  &  93.87  \\ 
 & \textsc{SUDS} \cite{turki2023suds} & RGB+LiDAR & 11.07  &  0.17  &  6.17  &  79.49  &  88.41  &  95.31  \\ 
 & \textsc{StreetSurf} \cite{guo2023streetsurf} & RGB+LiDAR & 10.86  &  0.15  &  5.90  &  82.16  &  93.57  &  96.96  \\ 
 & \textsc{LiDAR-NeRF} \cite{li2022practical} & LiDAR & 10.21  & 0.12 & \underline{ 4.72 } & \underline{ 87.71 } & \underline{ 95.05 } & \underline{ 97.78 } \\ 
 & \textbf{\textsc{Gated Fields}} \cite{li2022practical} & Gated & \textbf{7.92 } & \textbf{ 0.12 } & \textbf{ 4.13 } & \textbf{ 90.61 } & \textbf{ 95.76 } & \textbf{ 97.90 }

\\
			\midrule
			\multicolumn{9}{c}{\textbf{Test Data -- Day (Evaluated on Accumulated LiDAR Ground-Truth Points)}} \\
			\midrule
			\multirow{16}{*}{\rotatebox[origin=l]{90}{\parbox[c]{6cm}{\centering \textbf{\textsc{2D Depth Comparison}}}}}
            & \textsc{Gated2Gated}  \cite{gated2gated} & Gated & 9.26  &  0.22  &  6.69  &  58.46  &  93.70  &  97.38  \\ 
 & \textsc{GatedStereo} \cite{gatedstereo} & Stereo-Gated & \underline{6.32 } & 0.09 & \underline{ 3.30 } &  92.86  & \underline{ 97.31 } &  98.53  \\ 
 & \textsc{SimIPU} \cite{li2022simipu} & RGB & 13.54  &  0.31  &  10.08  &  52.90  &  86.49  &  95.65  \\ 
 & \textsc{AdaBins} \cite{bhat2021adabins} & RGB & 12.74  &  0.25  &  8.39  &  69.84  &  89.13  &  95.52  \\ 
 & \textsc{DPT} \cite{ranftl2021vision} & RGB & 10.34  &  0.21  &  7.08  &  77.61  &  94.29  &  97.24  \\ 
 & \textsc{DepthFormer} \cite{li2022depthformer} & RGB & 9.06  &  0.19  &  6.09  &  81.19  &  94.14  &  97.42  \\ 
 & \textsc{CREStereo} \cite{li2022practical} & Stereo-RGB & 7.35  & 0.09 &  3.45  & \textbf{ 94.48 } &  97.23  &  98.50  \\ 
 & \textsc{NLSPN} \cite{park2020nonRGBLidar} & RGB+LiDAR & 10.34  &  0.17  &  5.97  &  77.83  &  91.16  &  96.11  \\ 
 & \textsc{MipNeRF360} \cite{barron2022mipnerf360} & RGB & 16.91  &  0.31  &  9.53  &  70.89  &  84.43  &  90.92  \\ 
 & \textsc{K-Planes} \cite{fridovich2023kplanes} & RGB & 12.37  &  0.24  &  8.55  &  63.16  &  79.68  &  91.70  \\ 
 & \textsc{RawNeRF} \cite{mildenhall2022nerfdark} & RGB & 15.10  &  0.23  &  9.38  &  65.90  &  84.90  &  92.43  \\ 
 & \textsc{Depth-NeRF} \cite{deng2022depthNeRF} & RGB & 10.34  &  0.17  &  6.07  &  78.97  &  90.75  &  96.53  \\ 
 & \textsc{SUDS} \cite{turki2023suds} & RGB+LiDAR & 9.11  &  0.17  &  5.84  &  80.96  &  95.08  &  98.33  \\ 
 & \textsc{StreetSurf} \cite{guo2023streetsurf} & RGB+LiDAR & 9.60  &  0.13  &  5.35  &  83.94  &  95.32  & \underline{ 98.56 } \\ 
 & \textsc{LiDAR-NeRF} \cite{li2022practical} & LiDAR & 8.13  & \underline{ 0.10 } &  3.86  &  88.99  &  95.49  &  98.06  \\ 
 & \textbf{\textsc{Gated Fields}} \cite{li2022practical} & Gated & \textbf{6.15 } & \textbf{ 0.09 } & \textbf{ 2.91 } & \underline{ 93.88 } & \textbf{ 97.32 } & \textbf{ 98.75 }

\\
			\midrule
            \bottomrule
    \end{tabular}
    }
    \vspace*{-5pt}
    \caption{ Comparison of our proposed approach and state-of-the-art approaches on depth synthesis. Best results in each category are in \textbf{bold} and second best are \underline{underlined}.            }
    \label{tab:depth_res}
\end{table}

\begin{table}[!t]
	\centering
	\setlength{\tabcolsep}{2pt}
	\resizebox{0.89\linewidth}{!}{
		\begin{tabular}{l|cccccccccc}
			\toprule
			& \multirow{2}{*} \textbf{\textbf{Im. Formation}} &  \textbf{Depth} &   \textbf{Normal} &  \textbf{Active} &  \textbf{Shadow} &  \textbf{RMSE} & \textbf{MAE}  & $\textbf{PSNR}$ & \textbf{SSIM} \\ 
			& \textbf{Model} & \textbf{Sup.} &  & \textbf{Illum.} &  & $\left[ m \right]$ & $\left[ m \right]$ & $\left[ dB \right]$ & \\
			\midrule
		      \multicolumn{10}{c}{\textbf{Test Data -- Night }} \\
			\midrule
			\multirow{7}{*}{\rotatebox[origin=l]{90}{\parbox[c]{1.1cm}{\centering \textbf{\textsc{Ablation}}}}} 
			 & End2End \cite{mildenhall2020nerf} & \xmark & \xmark & \xmark & \xmark & 27.34  &  19.95  &  17.43  &  0.633    \\ 
 & Gated & \xmark & \xmark & \xmark & \xmark  & 9.34  &  8.35  &  23.47  &  0.889  \\ 
 & Gated & \cmark & \xmark & \xmark & \xmark & 6.72  &  5.43  &  25.84 &  0.908  \\ 
 & Gated & \cmark & \cmark & \cmark & \xmark & \underline{3.78}  &  \underline{2.82}  &  \underline{30.66}  &  \underline{0.946}  \\ 
 & Gated & \cmark & \cmark & \cmark & \cmark & \textbf{3.39}  &  \textbf{2.51}  &  \textbf{30.91}  &  \textbf{0.95}\\
			\midrule
    		\multicolumn{10}{c}{\textbf{Test Data -- Day }} \\
			\midrule
			\multirow{7}{*}{\rotatebox[origin=l]{90}{\parbox[c]{1.1cm}{\centering \textbf{\textsc{Ablation}}}}}
             & End2End \cite{mildenhall2020nerf} & \xmark & \xmark & \xmark & \xmark & 30.03  &  19.22 &  16.77  &  0.66    \\ 
 & Gated & \xmark & \xmark & \xmark & \xmark  & 11.64  &  6.34  & 26.77 &  0.915   \\ 
 & Gated & \cmark & \xmark & \xmark & \xmark & \underline{9.20}  &  \underline{4.22}  &  26.88  &  0.922   \\ 
 & Gated & \cmark & \cmark & \cmark & \xmark & 9.45 & 4.31 &  \underline{32.15}  &  \underline{0.946}  \\ 
 & Gated & \cmark & \cmark & \cmark & \cmark & \textbf{8.88}  &  \textbf{4.12} &  \textbf{32.28}  &  \textbf{0.948}\\
			\midrule
            \bottomrule
			
	\end{tabular}}
	\vspace*{-5pt}
	\caption{Ablation studies of the Gated Fields contributions, on a subset of the test dataset. We investigate different image formation models, neural fields components and supervision losses.}
	\label{tab:ablation}
	\vspace{-4mm}
\end{table}

\begin{figure*}[!t]
    \vspace{-5mm}
    \centering
    \noindent\includegraphics[width=1.0\textwidth]{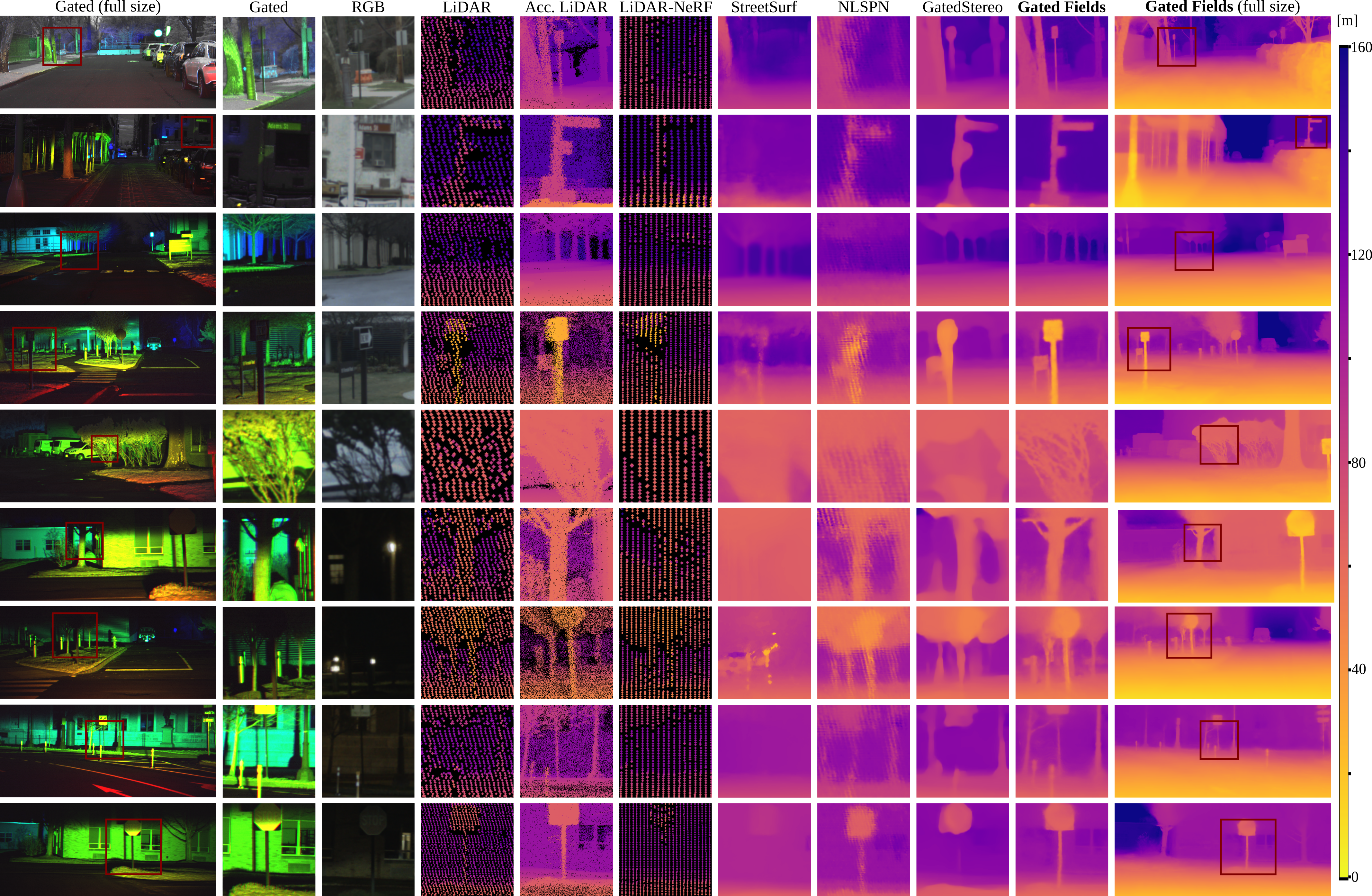} 
    \vspace{-3mm}
    \caption{Qualitative comparison of the proposed \textbf{Gated Fields} and state-of-the-art depth estimation approaches, including LiDAR-NeRF~\cite{tao2023lidarnerf}, StreetSurf \cite{guo2023streetsurf}, NLSPN \cite{park2020nonRGBLidar}, and Gated Stereo \cite{gatedstereo}. Compared to baseline methods, we are able to reconstruct fine geometry details like branches or poles, also for far distances. Unlike RGB methods, Gated Fields is unaffected by poor ambient lighting, and unlike LiDAR-based methods it is able to reconstruct sharp object discontinuities. The active gated slices are visualized in red for slice 1, green for slice 2 and blue for slice 3.}
    \label{fig:qual_results}
    \vspace{-4mm}
\end{figure*}

\begin{table}[!t]
    \vspace{-3mm}
	\centering
	\setlength{\tabcolsep}{4pt}
	\resizebox{0.82\linewidth}{!}{
		\begin{tabular}{l|cccccc}
			\toprule
			& \multirow{2}{*}{ \textbf{Method}} &  \textbf{Modality} &   \textbf{IoU} &  \textbf{Precision} &  \textbf{Recall}  \\ 
                &                                   &           &                 $\left[\%\right]$ & $\left[\%\right]$ & $\left[\%\right]$ \\
			\midrule
			\multirow{5}{*}{\rotatebox[origin=l]{90}{\parbox[c]{1.5cm}{\centering \small{ \textbf{\textsc{3D Rec.}}}}}} 
                & \textsc{MipNeRF360} \cite{barron2022mipnerf360} & RGB & 6.32 & 7.34 & 31.21\\    
                & \textsc{StreetSurf} \cite{guo2023streetsurf} & RGB+LiDAR & 5.41 &6.31 & 27.35   \\                 
                & \textsc{SUDS} \cite{turki2023suds} & RGB+LiDAR & 8.96 & 9.88 & 49.09\\ 
                & \textsc{LiDAR-NeRF} \cite{tao2023lidarnerf} & LiDAR &20.03 & \textbf{32.38} &34.44  \\ 
                & \textbf{Gated Fields} (ours) & Gated & \textbf{22.25} & \underline{25.01} & \textbf{66.51} \\ 
			\midrule
            \bottomrule
		\end{tabular}
	}
	\vspace*{-5pt}
	\caption{Comparison of Gated Fields and state-of-the-art scene reconstruction methods. We evaluate over 3D occupancy reconstruction, using as ground truth the voxelized accumulated LiDAR pointcloud.
 Best results in each category are in \textbf{bold} and second best are \underline{underlined}. }
	\label{tab:reconstruction}
	\vspace{-4mm}
\end{table}

\vspace{-12pt}
\paragraph{Depth Reconstruction}
We assess the quality of depth synthesis of Gated Fields for camera poses unseen during training. We use as ground truth the accumulated and filtered LiDAR pointcloud. Unlike previous works relying on single LiDAR scans for evaluation \cite{gated2depth2019,gated2gated,gatedstereo}, we use as ground truth an accumulated LiDAR pointcloud as described in \cref{dataset}. This allows us to evaluate the depth reconstruction up to \unit[160]{m} accurately and without bias. We follow previous works \cite{gated2gated,gatedstereo} and use as depth evaluation metrics RMSE, MAE, ARD, $\sigma_i < 1.25^i, i \in \{1,2,3 \}$. We compare our method against 9 feed-forward depth estimation methods, namely  SimIPU \cite{li2022simipu}, AdaBins \cite{bhat2021adabins}, DPT \cite{ranftl2021vision}, DepthFormer \cite{li2022depthformer} and CREStereo \cite{li2022practical} for monocular and stereo RGB methods, Gated2Gated \cite{gated2gated} and GatedStereo \cite{gatedstereo} for monocular and stereo gated estimation methods. We also compare our method against depths rendered with other neural reconstruction algorithms, using RGB images \cite{barron2022mipnerf360},\cite{mildenhall2022nerfdark},\cite{fridovich2023kplanes},\cite{deng2022depthNeRF} LiDAR \cite{tao2023lidarnerf}, RGB+LiDAR \cite{turki2023suds}, \cite{guo2023streetsurf}, and a varying-appearance method \cite{fridovich2023kplanes} for gated captures. Results for day and night sequences are presented in \cref{tab:depth_res}. We outperform the next best neural field method by 21.87$\%$ MAE and 30.35$\%$ in RMSE. For night sequences the performance difference sharpens, with Gated Fields outperforming the best RGB-based method \cite{li2022practical} by \unit[3.14]{m} MAE. This performance decline is to be attributed to the limited pixel information present in RGB captures taken at night time, making impossible to learn a meaningful 3D representation of the scene, as shown qualitatively in \cref{fig:qual_results}. LiDAR-based methods are unaffected by the change in illumination, but suffer from the limited sensor resolution. On the other hand, gated cameras retrieve information-rich captures both day and night, which Gated Fields can explicitly leverage during training. We confirm that employing state-of-the-art neural field methods on gated captures does not yield accurate results, as they are not able to model the gated imaging formation and can only fit the ambient light component.

\vspace{-16pt}
\paragraph{3D Reconstruction} 
We evaluate the 3D scene reconstruction capabilities of Gated Fields using the accumulated LiDAR pointcloud as ground truth. We follow \cite{cao2023scenerf} and extract for both the 3D ground truth pointcloud and different neural field-based methods a voxelized occupancy grid of the scene, and compute intersection over union (IoU), Precision and Recall between ground truth voxels and estimated ones. Quantitative results are shown in \cref{tab:reconstruction}. 
Our method outperforms RGB baselines \cite{barron2022mipnerf360,fridovich2023kplanes} by an average of 15\% IoU. RGB baselines using additional LiDAR sensor \cite{guo2023streetsurf,turki2023suds} data partially improve the results, but such methods are still unable to reconstruct finer surfaces details and struggle at night. On the other hand, Gated Fields is able to recover finer geometries using gated, illumination and depth cues, and the quality does not degrade with diminishing ambient light, as shown in \cref{fig:qual_results}. See details on evaluation and further qualitative results in the Supplementary Material.  

\vspace{-12pt}
\paragraph{Novel View Synthesis} 
For novel view synthesis, we compare our method with Mip-NeRF360 \cite{barron2022mipnerf360}, a state-of-the-art neural radiance field-based method, and K-Planes \cite{fridovich2023kplanes}, to implicitly model the time-varying appearance of the static scene. Mip-NeRF \cite{barron2022mipnerf360} struggles to reconstruct novel views due to the inherent difficulty of modeling the gating imaging effects, resulting in a PSNR of $17.16$dB. By learning a time-varying appearance, K-Planes improves the quality reaching $27.42$dB PSNR for day but only $19.35$dB for night, as the model fails to learn an accurate scene geometry representation without ambient light information. Gated Fields outperforms these baselines in both day and night, reaching a PSNR of $32.28$dB. 

\vspace{-12pt}
\paragraph{Ablation Experiments}
To assess the role and contribution of the different components of our method, we conduct an ablation study in Tab.~\ref{tab:ablation}. In particular, we consider as starting point a single neural field directly inferring one intensity value for each of the four slices (3 active + 1 passive), and obtain an average MAE of \unit[19.58]{m}. By separately predicting ambient light and reflectance, and reconstructing the gated image as in \cref{eq:toteq_gated_basic}, we significantly improve the MAE to \unit[7.34]{m} . However, this approach still performs poorly on flat-color areas during the day and in un-illuminated areas during the night due to lack of any depth cue. By adding the depth supervision, we are able to supervise also such areas and the PSNR improves by $4.83$dB. By adding the angular-dependent attenuation and regularizing the reflectance in \cref{eq:reflectance_reg_loss}, the model is able to disentangle the material proprieties from other spurious effects. Finally, by explicitly modeling the shadow, casted by the illuminator, we improve final depth reconstruction to \unit[3.32]{m} MAE. 

\section{Conclusion}
We introduce Gated Fields, a neural rendering method capable of reconstructing scene geometry from video captures of active time-gated cameras. The method hinges on a differentiable gated image formation as part of the rendering formulation, and it jointly learns geometry, ambient light and surface proprieties, represented implicitly as neural field components, alongside illumination and gating parameters, represented with physics-based models. Extensive experiments on real-world large-scale scenes validate that our method is able to precisely reconstruct a 3D scene both in day and night-time conditions. Our approach outperforms existing RGB and LiDAR methods by $21.87\%$ on MAE, as well as baseline methods using gated captures by $31.67\%$. In the future, we hope to extend the proposed approach by ``closing the loop'' and providing dynamic feedback to the gated acquisition, allowing for adaptive gated scene reconstructions.

\vspace{0.2\baselineskip}\PAR{Acknowledgments} This work was supported by the AI-SEE project with funding from the FFG, BMBF, and NRC-IRA. Felix Heide was supported by an NSF CAREER Award (2047359), a Packard Foundation Fellowship, a Sloan Research Fellowship, a Sony Young Faculty Award, a Project X Innovation Award, and an Amazon Science Research Award.

\newpage

\newpage
{
    \small
    \bibliographystyle{ieeenat_fullname}
    \bibliography{main}

\begin{thebibliography}{103}
\providecommand{\natexlab}[1]{#1}
\providecommand{\url}[1]{\texttt{#1}}
\expandafter\ifx\csname urlstyle\endcsname\relax
  \providecommand{\doi}[1]{doi: #1}\else
  \providecommand{\doi}{doi: \begingroup \urlstyle{rm}\Url}\fi

\bibitem[Adam et~al.(2017)Adam, Dann, Yair, Mazor, and
  Nowozin]{adam2017bayesian}
Amit Adam, Christoph Dann, Omer Yair, Shai Mazor, and Sebastian Nowozin.
\newblock Bayesian time-of-flight for realtime shape, illumination and albedo.
\newblock 39\penalty0 (5):\penalty0 851--864, 2017.

\bibitem[Agarwal and Arora(2023)]{agarwal2023attention}
Ashutosh Agarwal and Chetan Arora.
\newblock Attention attention everywhere: Monocular depth prediction with skip
  attention.
\newblock In \emph{Proceedings of the IEEE/CVF Winter Conference on
  Applications of Computer Vision}, pages 5861--5870, 2023.

\bibitem[Andersson(2006)]{Andersson2006}
Pierre Andersson.
\newblock Long-range three-dimensional imaging using range-gated laser radar
  images.
\newblock 45\penalty0 (3):\penalty0 034301, 2006.

\bibitem[Attal et~al.(2021)Attal, Laidlaw, Gokaslan, Kim, Richardt, Tompkin,
  and O'Toole]{attal2021torf}
Benjamin Attal, Eliot Laidlaw, Aaron Gokaslan, Changil Kim, Christian Richardt,
  James Tompkin, and Matthew O'Toole.
\newblock T{\"o}rf: Time-of-flight radiance fields for dynamic scene view
  synthesis.
\newblock \emph{Advances in neural information processing systems},
  34:\penalty0 26289--26301, 2021.

\bibitem[Badki et~al.(2020)Badki, Troccoli, Kim, Kautz, Sen, and
  Gallo]{badki2020Bi3D}
Abhishek Badki, Alejandro Troccoli, Kihwan Kim, Jan Kautz, Pradeep Sen, and
  Orazio Gallo.
\newblock {Bi3D}: {S}tereo depth estimation via binary classifications.
\newblock In \emph{arXiv preprint arXiv:2005.07274}, 2020.

\bibitem[Barron et~al.(2021)Barron, Mildenhall, Tancik, Hedman, Martin-Brualla,
  and Srinivasan]{barron2021mipnerf}
Jonathan~T. Barron, Ben Mildenhall, Matthew Tancik, Peter Hedman, Ricardo
  Martin-Brualla, and Pratul~P. Srinivasan.
\newblock Mip-nerf: A multiscale representation for anti-aliasing neural
  radiance fields.
\newblock \emph{ICCV}, 2021.

\bibitem[Barron et~al.(2022)Barron, Mildenhall, Verbin, Srinivasan, and
  Hedman]{barron2022mipnerf360}
Jonathan~T Barron, Ben Mildenhall, Dor Verbin, Pratul~P Srinivasan, and Peter
  Hedman.
\newblock Mip-nerf 360: Unbounded anti-aliased neural radiance fields.
\newblock In \emph{Proceedings of the IEEE/CVF Conference on Computer Vision
  and Pattern Recognition}, pages 5470--5479, 2022.

\bibitem[Barron et~al.(2023)Barron, Mildenhall, Verbin, Srinivasan, and
  Hedman]{barron2023zipnerf}
Jonathan~T. Barron, Ben Mildenhall, Dor Verbin, Pratul~P. Srinivasan, and Peter
  Hedman.
\newblock Zip-nerf: Anti-aliased grid-based neural radiance fields.
\newblock \emph{ICCV}, 2023.

\bibitem[Bartoccioni et~al.(2023)Bartoccioni, Zablocki, P{\'e}rez, Cord, and
  Alahari]{bartoccioni2023lidartouch}
Florent Bartoccioni, {\'E}loi Zablocki, Patrick P{\'e}rez, Matthieu Cord, and
  Karteek Alahari.
\newblock Lidartouch: Monocular metric depth estimation with a few-beam lidar.
\newblock \emph{Computer Vision and Image Understanding}, 227:\penalty0 103601,
  2023.

\bibitem[Bhat et~al.(2021)Bhat, Alhashim, and Wonka]{bhat2021adabins}
Shariq~Farooq Bhat, Ibraheem Alhashim, and Peter Wonka.
\newblock Adabins: Depth estimation using adaptive bins.
\newblock In \emph{Proceedings of the IEEE/CVF Conference on Computer Vision
  and Pattern Recognition}, pages 4009--4018, 2021.

\bibitem[Bijelic et~al.(2018{\natexlab{a}})Bijelic, Gruber, and
  Ritter]{BenchmarkLidar}
Mario Bijelic, Tobias Gruber, and Werner Ritter.
\newblock A benchmark for lidar sensors in fog: Is detection breaking down?
\newblock In \emph{2018 IEEE Intelligent Vehicles Symposium (IV)}, pages
  760--767, 2018{\natexlab{a}}.

\bibitem[Bijelic et~al.(2018{\natexlab{b}})Bijelic, Gruber, and
  Ritter]{Bijelic2018}
Mario Bijelic, Tobias Gruber, and Werner Ritter.
\newblock Benchmarking image sensors under adverse weather conditions for
  autonomous driving.
\newblock 2018{\natexlab{b}}.

\bibitem[Bijelic et~al.(2018{\natexlab{c}})Bijelic, Gruber, and
  Ritter]{bijelic2018benchmarking}
Mario Bijelic, Tobias Gruber, and Werner Ritter.
\newblock Benchmarking image sensors under adverse weather conditions for
  autonomous driving.
\newblock In \emph{2018 IEEE Intelligent Vehicles Symposium (IV)}, pages
  1773--1779. IEEE, 2018{\natexlab{c}}.

\bibitem[Bijelic et~al.(2020)Bijelic, Gruber, Mannan, Kraus, Ritter, Dietmayer,
  and Heide]{Bijelic_2020_STF}
Mario Bijelic, Tobias Gruber, Fahim Mannan, Florian Kraus, Werner Ritter, Klaus
  Dietmayer, and Felix Heide.
\newblock Seeing through fog without seeing fog: Deep multimodal sensor fusion
  in unseen adverse weather.
\newblock In \emph{The IEEE Conference on Computer Vision and Pattern
  Recognition (CVPR)}, 2020.

\bibitem[Boss et~al.(2022)Boss, Engelhardt, Kar, Li, Sun, Barron, Lensch, and
  Jampani]{boss2022samurai}
Mark Boss, Andreas Engelhardt, Abhishek Kar, Yuanzhen Li, Deqing Sun, Jonathan
  Barron, Hendrik Lensch, and Varun Jampani.
\newblock Samurai: Shape and material from unconstrained real-world arbitrary
  image collections.
\newblock \emph{Advances in Neural Information Processing Systems},
  35:\penalty0 26389--26403, 2022.

\bibitem[Bozic et~al.(2021)Bozic, Palafox, Thies, Dai, and
  Nie{\ss}ner]{bozic2021transformerfusion}
Aljaz Bozic, Pablo Palafox, Justus Thies, Angela Dai, and Matthias Nie{\ss}ner.
\newblock Transformerfusion: Monocular rgb scene reconstruction using
  transformers.
\newblock \emph{Advances in Neural Information Processing Systems},
  34:\penalty0 1403--1414, 2021.

\bibitem[Busck(2005)]{Busck2005}
Jens Busck.
\newblock Underwater {3-D} optical imaging with a gated viewing laser radar.
\newblock 2005.

\bibitem[Busck and Heiselberg(2004{\natexlab{a}})]{Busck2004}
Jens Busck and Henning Heiselberg.
\newblock Gated viewing and high-accuracy three-dimensional laser radar.
\newblock 43\penalty0 (24):\penalty0 4705--10, 2004{\natexlab{a}}.

\bibitem[Busck and Heiselberg(2004{\natexlab{b}})]{busck2004gated}
Jens Busck and Henning Heiselberg.
\newblock Gated viewing and high-accuracy three-dimensional laser radar.
\newblock \emph{Applied optics}, 43\penalty0 (24):\penalty0 4705--4710,
  2004{\natexlab{b}}.

\bibitem[Cao and de~Charette(2023)]{cao2023scenerf}
Anh-Quan Cao and Raoul de Charette.
\newblock Scenerf: Self-supervised monocular 3d scene reconstruction with
  radiance fields.
\newblock In \emph{Proceedings of the IEEE/CVF International Conference on
  Computer Vision}, pages 9387--9398, 2023.

\bibitem[{Carballo} et~al.(2020){Carballo}, {Lambert}, {Monrroy}, {Wong},
  {Narksri}, {Kitsukawa}, {Takeuchi}, {Kato}, and {Takeda}]{LIBRE}
A. {Carballo}, J. {Lambert}, A. {Monrroy}, D. {Wong}, P. {Narksri}, Y.
  {Kitsukawa}, E. {Takeuchi}, S. {Kato}, and K. {Takeda}.
\newblock Libre: The multiple 3d lidar dataset.
\newblock In \emph{IEEE Intelligent Vehicles Symposium (IV)}, 2020.

\bibitem[Chang and Chen(2018)]{chang2018pyramid}
Jia-Ren Chang and Yong-Sheng Chen.
\newblock Pyramid stereo matching network.
\newblock In \emph{Proceedings of the IEEE Conference on Computer Vision and
  Pattern Recognition}, pages 5410--5418, 2018.

\bibitem[Chen et~al.(2022)Chen, Xu, Geiger, Yu, and Su]{chen2022tensorf}
Anpei Chen, Zexiang Xu, Andreas Geiger, Jingyi Yu, and Hao Su.
\newblock Tensorf: Tensorial radiance fields.
\newblock In \emph{European Conference on Computer Vision}, pages 333--350.
  Springer, 2022.

\bibitem[Choe et~al.(2021)Choe, Joo, Imtiaz, and
  Kweon]{VolPropagationNetStereoLidar}
Jaesung Choe, Kyungdon Joo, Tooba Imtiaz, and In~So Kweon.
\newblock Volumetric propagation network: Stereo-lidar fusion for long-range
  depth estimation.
\newblock \emph{IEEE Robotics and Automation Letters}, 6\penalty0 (3):\penalty0
  4672--4679, 2021.

\bibitem[Curless and Levoy(1996)]{curless1996tsdf}
Brian Curless and Marc Levoy.
\newblock A volumetric method for building complex models from range images.
\newblock In \emph{Proceedings of the 23rd annual conference on Computer
  graphics and interactive techniques}, pages 303--312, 1996.

\bibitem[Deng et~al.(2022)Deng, Liu, Zhu, and Ramanan]{deng2022depthNeRF}
Kangle Deng, Andrew Liu, Jun-Yan Zhu, and Deva Ramanan.
\newblock Depth-supervised nerf: Fewer views and faster training for free.
\newblock In \emph{Proceedings of the IEEE/CVF Conference on Computer Vision
  and Pattern Recognition}, pages 12882--12891, 2022.

\bibitem[Eigen et~al.(2014)Eigen, Puhrsch, and Fergus]{eigen2014depth}
David Eigen, Christian Puhrsch, and Rob Fergus.
\newblock Depth map prediction from a single image using a multi-scale deep
  network.
\newblock In \emph{NeurIPS}, pages 2366--2374, 2014.

\bibitem[Fridovich-Keil et~al.(2022)Fridovich-Keil, Yu, Tancik, Chen, Recht,
  and Kanazawa]{fridovich2022plenoxels}
Sara Fridovich-Keil, Alex Yu, Matthew Tancik, Qinhong Chen, Benjamin Recht, and
  Angjoo Kanazawa.
\newblock Plenoxels: Radiance fields without neural networks.
\newblock In \emph{Proceedings of the IEEE/CVF Conference on Computer Vision
  and Pattern Recognition}, pages 5501--5510, 2022.

\bibitem[Fridovich-Keil et~al.(2023)Fridovich-Keil, Meanti, Warburg, Recht, and
  Kanazawa]{fridovich2023kplanes}
Sara Fridovich-Keil, Giacomo Meanti, Frederik~Rahb{\ae}k Warburg, Benjamin
  Recht, and Angjoo Kanazawa.
\newblock K-planes: Explicit radiance fields in space, time, and appearance.
\newblock In \emph{Proceedings of the IEEE/CVF Conference on Computer Vision
  and Pattern Recognition}, pages 12479--12488, 2023.

\bibitem[Garg et~al.(2016)Garg, Kumar, Carneiro, and Reid]{Garg2016}
Ravi Garg, B.G.~Vijay Kumar, Gustavo Carneiro, and Ian Reid.
\newblock Unsupervised {CNN} for single view depth estimation: Geometry to the
  rescue.
\newblock In \emph{ECCV}, pages 740--756, 2016.

\bibitem[Geiger et~al.(2012)Geiger, Lenz, and Urtasun]{geiger2012we}
Andreas Geiger, Philip Lenz, and Raquel Urtasun.
\newblock Are we ready for autonomous driving? the kitti vision benchmark
  suite.
\newblock In \emph{CVPR}, pages 3354--3361, 2012.

\bibitem[Godard et~al.(2017)Godard, Mac~Aodha, and
  Brostow]{godard2017unsupervised}
Cl{\'e}ment Godard, Oisin Mac~Aodha, and Gabriel~J Brostow.
\newblock Unsupervised monocular depth estimation with left-right consistency.
\newblock In \emph{Proceedings of the IEEE conference on computer vision and
  pattern recognition}, pages 270--279, 2017.

\bibitem[Godard et~al.(2019)Godard, Mac~Aodha, Firman, and
  Brostow]{godard2019digging}
Cl{\'e}ment Godard, Oisin Mac~Aodha, Michael Firman, and Gabriel~J Brostow.
\newblock Digging into self-supervised monocular depth estimation.
\newblock In \emph{Proceedings of the IEEE/CVF International Conference on
  Computer Vision}, pages 3828--3838, 2019.

\bibitem[Grauer(2014)]{grauer2014active}
Yoav Grauer.
\newblock Active gated imaging in driver assistance system.
\newblock \emph{Advanced Optical Technologies}, 3\penalty0 (2):\penalty0
  151--160, 2014.

\bibitem[Gruber et~al.(2019)Gruber, Julca-Aguilar, Bijelic, and
  Heide]{gated2depth2019}
Tobias Gruber, Frank Julca-Aguilar, Mario Bijelic, and Felix Heide.
\newblock Gated2depth: Real-time dense lidar from gated images.
\newblock In \emph{The IEEE International Conference on Computer Vision
  (ICCV)}, 2019.

\bibitem[Guizilini et~al.(2020)Guizilini, Ambrus, Pillai, Raventos, and
  Gaidon]{guizilini20203d}
Vitor Guizilini, Rares Ambrus, Sudeep Pillai, Allan Raventos, and Adrien
  Gaidon.
\newblock 3d packing for self-supervised monocular depth estimation.
\newblock In \emph{Proceedings of the IEEE/CVF Conference on Computer Vision
  and Pattern Recognition}, pages 2485--2494, 2020.

\bibitem[Guo et~al.(2023)Guo, Deng, Li, Bai, Shi, Wang, Ding, Wang, and
  Li]{guo2023streetsurf}
Jianfei Guo, Nianchen Deng, Xinyang Li, Yeqi Bai, Botian Shi, Chiyu Wang,
  Chenjing Ding, Dongliang Wang, and Yikang Li.
\newblock Streetsurf: Extending multi-view implicit surface reconstruction to
  street views.
\newblock \emph{arXiv preprint arXiv:2306.04988}, 2023.

\bibitem[Hansard et~al.(2012)Hansard, Lee, Choi, and Horaud]{hansard2012time}
Miles Hansard, Seungkyu Lee, Ouk Choi, and Radu~Patrice Horaud.
\newblock \emph{Time-of-flight cameras: principles, methods and applications}.
\newblock Springer Science \& Business Media, 2012.

\bibitem[Heckman and Hodgson(1967)]{heckman1967}
Paul Heckman and Robert~T. Hodgson.
\newblock Underwater optical range gating.
\newblock 3\penalty0 (11):\penalty0 445--448, 1967.

\bibitem[Hu et~al.(2021)Hu, Wang, Li, Ning, Fan, and Gong]{hu2020PENetRGBLidar}
Mu Hu, Shuling Wang, Bin Li, Shiyu Ning, Li Fan, and Xiaojin Gong.
\newblock Towards precise and efficient image guided depth completion.
\newblock 2021.

\bibitem[Huang et~al.(2023)Huang, Gojcic, Wang, Williams, Kasten, Fidler,
  Schindler, and Litany]{huang2023nfl}
Shengyu Huang, Zan Gojcic, Zian Wang, Francis Williams, Yoni Kasten, Sanja
  Fidler, Konrad Schindler, and Or Litany.
\newblock Neural lidar fields for novel view synthesis.
\newblock 2023.

\bibitem[Jaritz et~al.(2018)Jaritz, De~Charette, Wirbel, Perrotton, and
  Nashashibi]{jaritz2018sparse}
Maximilian Jaritz, Raoul De~Charette, Emilie Wirbel, Xavier Perrotton, and
  Fawzi Nashashibi.
\newblock Sparse and dense data with cnns: Depth completion and semantic
  segmentation.
\newblock pages 52--60, 2018.

\bibitem[Jokela et~al.(2019)Jokela, Kutila, and Pyykönen]{Jokela}
Maria Jokela, Matti Kutila, and Pasi Pyykönen.
\newblock Testing and validation of automotive point-cloud sensors in adverse
  weather conditions.
\newblock \emph{Applied Sciences}, 9, 2019.

\bibitem[Julca-Aguilar et~al.(2021)Julca-Aguilar, Taylor, Bijelic, Mannan,
  Tseng, and Heide]{julca2021gated3d}
Frank Julca-Aguilar, Jason Taylor, Mario Bijelic, Fahim Mannan, Ethan Tseng,
  and Felix Heide.
\newblock Gated3d: Monocular 3d object detection from temporal illumination
  cues.
\newblock In \emph{Proceedings of the IEEE/CVF International Conference on
  Computer Vision}, pages 2938--2948, 2021.

\bibitem[Kendall et~al.(2017)Kendall, Martirosyan, Dasgupta, Henry, Kennedy,
  Bachrach, and Bry]{Kendall2017}
Alex Kendall, Hayk Martirosyan, Saumitro Dasgupta, Peter Henry, Ryan Kennedy,
  Abraham Bachrach, and Adam Bry.
\newblock End-to-end learning of geometry and context for deep stereo
  regression.
\newblock In \emph{ICCV}, 2017.

\bibitem[Kolb et~al.(2010)Kolb, Barth, Koch, and Larsen]{kolb2010time}
Andreas Kolb, Erhardt Barth, Reinhard Koch, and Rasmus Larsen.
\newblock Time-of-flight cameras in computer graphics.
\newblock In \emph{Computer Graphics Forum}, pages 141--159. Wiley Online
  Library, 2010.

\bibitem[Kundu et~al.(2022)Kundu, Genova, Yin, Fathi, Pantofaru, Guibas,
  Tagliasacchi, Dellaert, and Funkhouser]{kundu2022panoptic}
Abhijit Kundu, Kyle Genova, Xiaoqi Yin, Alireza Fathi, Caroline Pantofaru,
  Leonidas~J Guibas, Andrea Tagliasacchi, Frank Dellaert, and Thomas
  Funkhouser.
\newblock Panoptic neural fields: A semantic object-aware neural scene
  representation.
\newblock In \emph{Proceedings of the IEEE/CVF Conference on Computer Vision
  and Pattern Recognition}, pages 12871--12881, 2022.

\bibitem[Lange(2000)]{lange00tof}
Robert Lange.
\newblock {3D} time-of-flight distance measurement with custom solid-state
  image sensors in {CMOS}/{CCD}-technology.
\newblock 2000.

\bibitem[Levy et~al.(2023)Levy, Peleg, Pearl, Rosenbaum, Akkaynak, Korman, and
  Treibitz]{levy2023seathru}
Deborah Levy, Amit Peleg, Naama Pearl, Dan Rosenbaum, Derya Akkaynak, Simon
  Korman, and Tali Treibitz.
\newblock Seathru-nerf: Neural radiance fields in scattering media.
\newblock In \emph{Proceedings of the IEEE/CVF Conference on Computer Vision
  and Pattern Recognition}, pages 56--65, 2023.

\bibitem[Li et~al.(2022{\natexlab{a}})Li, Wang, Xiong, Cai, Yan, Yang, Liu,
  Fan, and Liu]{li2022practical}
Jiankun Li, Peisen Wang, Pengfei Xiong, Tao Cai, Ziwei Yan, Lei Yang, Jiangyu
  Liu, Haoqiang Fan, and Shuaicheng Liu.
\newblock Practical stereo matching via cascaded recurrent network with
  adaptive correlation.
\newblock In \emph{Proceedings of the IEEE/CVF conference on computer vision
  and pattern recognition}, pages 16263--16272, 2022{\natexlab{a}}.

\bibitem[Li et~al.(2021{\natexlab{a}})Li, Dekel, Cole, Tucker, Snavely, Liu,
  and Freeman]{MovingPeopleMovingCameras}
Zhengqi Li, Tali Dekel, Forrester Cole, Richard Tucker, Noah Snavely, Ce Liu,
  and William~T. Freeman.
\newblock Mannequinchallenge: Learning the depths of moving people by watching
  frozen people.
\newblock \emph{IEEE Transactions on Pattern Analysis and Machine
  Intelligence}, 43\penalty0 (12):\penalty0 4229--4241, 2021{\natexlab{a}}.

\bibitem[Li et~al.(2021{\natexlab{b}})Li, Liu, Drenkow, Ding, Creighton,
  Taylor, and Unberath]{TransformerStereo}
Zhaoshuo Li, Xingtong Liu, Nathan Drenkow, Andy Ding, Francis~X. Creighton,
  Russell~H. Taylor, and Mathias Unberath.
\newblock Revisiting stereo depth estimation from a sequence-to-sequence
  perspective with transformers.
\newblock In \emph{2021 IEEE/CVF International Conference on Computer Vision
  (ICCV)}, pages 6177--6186, 2021{\natexlab{b}}.

\bibitem[Li et~al.(2022{\natexlab{b}})Li, Chen, Li, Fang, Jiang, Liu, Jiang,
  Zhou, and Zhao]{li2022simipu}
Zhenyu Li, Zehui Chen, Ang Li, Liangji Fang, Qinhong Jiang, Xianming Liu,
  Junjun Jiang, Bolei Zhou, and Hang Zhao.
\newblock Simipu: Simple 2d image and 3d point cloud unsupervised pre-training
  for spatial-aware visual representations.
\newblock In \emph{Proceedings of the AAAI Conference on Artificial
  Intelligence}, pages 1500--1508, 2022{\natexlab{b}}.

\bibitem[Li et~al.(2022{\natexlab{c}})Li, Chen, Liu, and
  Jiang]{li2022depthformer}
Zhenyu Li, Zehui Chen, Xianming Liu, and Junjun Jiang.
\newblock Depthformer: Depthformer: Exploiting long-range correlation and local
  information for accurate monocular depth estimation.
\newblock \emph{arXiv preprint arXiv:2203.14211}, 2022{\natexlab{c}}.

\bibitem[Li et~al.(2023)Li, M\"uller, Evans, Taylor, Unberath, Liu, and
  Lin]{li2023neuralangelo}
Zhaoshuo Li, Thomas M\"uller, Alex Evans, Russell~H Taylor, Mathias Unberath,
  Ming-Yu Liu, and Chen-Hsuan Lin.
\newblock Neuralangelo: High-fidelity neural surface reconstruction.
\newblock In \emph{IEEE Conference on Computer Vision and Pattern Recognition
  ({CVPR})}, 2023.

\bibitem[Liu et~al.(2023)Liu, Chen, Yang, Wang, Manivasagam, and
  Urtasun]{liu2023waabiICCV23}
Jeffrey~Yunfan Liu, Yun Chen, Ze Yang, Jingkang Wang, Sivabalan Manivasagam,
  and Raquel Urtasun.
\newblock Real-time neural rasterization for large scenes.
\newblock In \emph{Proceedings of the IEEE/CVF International Conference on
  Computer Vision}, pages 8416--8427, 2023.

\bibitem[Loshchilov and Hutter(2018)]{ADAMW}
Ilya Loshchilov and Frank Hutter.
\newblock Decoupled weight decay regularization.
\newblock \emph{International Conference on Learning Representations}, 2018.

\bibitem[Ma and Karaman(2018)]{ma2018sparse}
Fangchang Ma and Sertac Karaman.
\newblock Sparse-to-dense: Depth prediction from sparse depth samples and a
  single image.
\newblock pages 1--8, 2018.

\bibitem[Malik et~al.(2023)Malik, Mirdehghan, Nousias, Kutulakos, and
  Lindell]{malik2023transientnerf}
Anagh Malik, Parsa Mirdehghan, Sotiris Nousias, Kiriakos~N Kutulakos, and
  David~B Lindell.
\newblock Transient neural radiance fields for lidar view synthesis and 3d
  reconstruction.
\newblock \emph{arXiv preprint arXiv:2307.09555}, 2023.

\bibitem[Mayer et~al.(2016)Mayer, Ilg, H{\"a}usser, Fischer, Cremers,
  Dosovitskiy, and Brox]{Mayer2016}
N. Mayer, E. Ilg, P. H{\"a}usser, P. Fischer, D. Cremers, A. Dosovitskiy, and
  T. Brox.
\newblock A large dataset to train convolutional networks for disparity,
  optical flow, and scene flow estimation.
\newblock In \emph{CVPR}, 2016.

\bibitem[Meuleman et~al.(2023)Meuleman, Liu, Gao, Huang, Kim, Kim, and
  Kopf]{meuleman2023localrf}
Andreas Meuleman, Yu-Lun Liu, Chen Gao, Jia-Bin Huang, Changil Kim, Min~H. Kim,
  and Johannes Kopf.
\newblock Progressively optimized local radiance fields for robust view
  synthesis.
\newblock In \emph{CVPR}, 2023.

\bibitem[Mildenhall et~al.(2020)Mildenhall, Srinivasan, Tancik, Barron,
  Ramamoorthi, and Ng]{mildenhall2020nerf}
Ben Mildenhall, Pratul~P. Srinivasan, Matthew Tancik, Jonathan~T. Barron, Ravi
  Ramamoorthi, and Ren Ng.
\newblock Nerf: Representing scenes as neural radiance fields for view
  synthesis.
\newblock In \emph{ECCV}, 2020.

\bibitem[Mildenhall et~al.(2022)Mildenhall, Hedman, Martin-Brualla, Srinivasan,
  and Barron]{mildenhall2022nerfdark}
Ben Mildenhall, Peter Hedman, Ricardo Martin-Brualla, Pratul~P Srinivasan, and
  Jonathan~T Barron.
\newblock Nerf in the dark: High dynamic range view synthesis from noisy raw
  images.
\newblock In \emph{Proceedings of the IEEE/CVF Conference on Computer Vision
  and Pattern Recognition}, pages 16190--16199, 2022.

\bibitem[M{\"u}ller et~al.(2022)M{\"u}ller, Evans, Schied, and
  Keller]{muller2022instantNGP}
Thomas M{\"u}ller, Alex Evans, Christoph Schied, and Alexander Keller.
\newblock Instant neural graphics primitives with a multiresolution hash
  encoding.
\newblock \emph{ACM Transactions on Graphics (ToG)}, 41\penalty0 (4):\penalty0
  1--15, 2022.

\bibitem[Murez et~al.(2020)Murez, Van~As, Bartolozzi, Sinha, Badrinarayanan,
  and Rabinovich]{murez2020atlas}
Zak Murez, Tarrence Van~As, James Bartolozzi, Ayan Sinha, Vijay Badrinarayanan,
  and Andrew Rabinovich.
\newblock Atlas: End-to-end 3d scene reconstruction from posed images.
\newblock In \emph{Computer Vision--ECCV 2020: 16th European Conference,
  Glasgow, UK, August 23--28, 2020, Proceedings, Part VII 16}, pages 414--431.
  Springer, 2020.

\bibitem[Ost et~al.(2022)Ost, Laradji, Newell, Bahat, and
  Heide]{ost2022pointlightfields}
Julian Ost, Issam Laradji, Alejandro Newell, Yuval Bahat, and Felix Heide.
\newblock Neural point light fields.
\newblock \emph{Proceedings of the IEEE Conference on Computer Vision and
  Pattern Recognition (CVPR)}, 2022.

\bibitem[Park et~al.(2020)Park, Joo, Hu, Liu, and Kweon]{park2020nonRGBLidar}
Jinsun Park, Kyungdon Joo, Zhe Hu, Chi-Kuei Liu, and In~So Kweon.
\newblock Non-local spatial propagation network for depth completion.
\newblock In \emph{Proc. of European Conference on Computer Vision (ECCV)},
  2020.

\bibitem[Pradeep et~al.(2013)Pradeep, Rhemann, Izadi, Zach, Bleyer, and
  Bathiche]{pradeep2013monofusion}
Vivek Pradeep, Christoph Rhemann, Shahram Izadi, Christopher Zach, Michael
  Bleyer, and Steven Bathiche.
\newblock Monofusion: Real-time 3d reconstruction of small scenes with a single
  web camera.
\newblock In \emph{2013 IEEE International Symposium on Mixed and Augmented
  Reality (ISMAR)}, pages 83--88. IEEE, 2013.

\bibitem[Ramazzina et~al.(2023)Ramazzina, Bijelic, Walz, Sanvito, Scheuble, and
  Heide]{ramazzina2023scatternerf}
Andrea Ramazzina, Mario Bijelic, Stefanie Walz, Alessandro Sanvito, Dominik
  Scheuble, and Felix Heide.
\newblock Scatternerf: Seeing through fog with physically-based inverse neural
  rendering.
\newblock \emph{arXiv preprint arXiv:2305.02103}, 2023.

\bibitem[Ranftl et~al.(2021)Ranftl, Bochkovskiy, and Koltun]{ranftl2021vision}
Ren{\'e} Ranftl, Alexey Bochkovskiy, and Vladlen Koltun.
\newblock Vision transformers for dense prediction.
\newblock In \emph{Proceedings of the IEEE/CVF International Conference on
  Computer Vision}, pages 12179--12188, 2021.

\bibitem[Rematas et~al.(2022)Rematas, Liu, Srinivasan, Barron, Tagliasacchi,
  Funkhouser, and Ferrari]{rematas2022urf}
Konstantinos Rematas, Andrew Liu, Pratul~P. Srinivasan, Jonathan~T. Barron,
  Andrea Tagliasacchi, Tom Funkhouser, and Vittorio Ferrari.
\newblock Urban radiance fields.
\newblock \emph{CVPR}, 2022.

\bibitem[Riegler et~al.(2017)Riegler, Ulusoy, Bischof, and
  Geiger]{riegler2017octnetfusion}
Gernot Riegler, Ali~Osman Ulusoy, Horst Bischof, and Andreas Geiger.
\newblock Octnetfusion: Learning depth fusion from data.
\newblock In \emph{2017 International Conference on 3D Vision (3DV)}, pages
  57--66. IEEE, 2017.

\bibitem[Roessle et~al.(2022)Roessle, Barron, Mildenhall, Srinivasan, and
  Nie{\ss}ner]{roessle2022densedepthNeRF}
Barbara Roessle, Jonathan~T Barron, Ben Mildenhall, Pratul~P Srinivasan, and
  Matthias Nie{\ss}ner.
\newblock Dense depth priors for neural radiance fields from sparse input
  views.
\newblock In \emph{Proceedings of the IEEE/CVF Conference on Computer Vision
  and Pattern Recognition}, pages 12892--12901, 2022.

\bibitem[Rudnev et~al.(2022)Rudnev, Elgharib, Smith, Liu, Golyanik, and
  Theobalt]{rudnev2022nerfOSR}
Viktor Rudnev, Mohamed Elgharib, William Smith, Lingjie Liu, Vladislav
  Golyanik, and Christian Theobalt.
\newblock Nerf for outdoor scene relighting.
\newblock In \emph{European Conference on Computer Vision}, pages 615--631.
  Springer, 2022.

\bibitem[Schober et~al.(2017)Schober, Adam, Yair, Mazor, and
  Nowozin]{schober2017dynamic}
Michael Schober, Amit Adam, Omer Yair, Shai Mazor, and Sebastian Nowozin.
\newblock Dynamic time-of-flight.
\newblock In \emph{CVPR}, pages 6109--6118, 2017.

\bibitem[Sch{\"o}ps et~al.(2015)Sch{\"o}ps, Sattler, H{\"a}ne, and
  Pollefeys]{schops20153d}
Thomas Sch{\"o}ps, Torsten Sattler, Christian H{\"a}ne, and Marc Pollefeys.
\newblock 3d modeling on the go: Interactive 3d reconstruction of large-scale
  scenes on mobile devices.
\newblock In \emph{2015 International Conference on 3D Vision}, pages 291--299.
  IEEE, 2015.

\bibitem[Schwarz(2010)]{schwarz2010lidar}
Brent Schwarz.
\newblock Lidar: Mapping the world in {3D}.
\newblock \emph{Nature Photonics}, 4\penalty0 (7):\penalty0 429, 2010.

\bibitem[Shan et~al.(2020)Shan, Englot, Meyers, Wang, Ratti, and
  Rus]{shan2020liosam}
Tixiao Shan, Brendan Englot, Drew Meyers, Wei Wang, Carlo Ratti, and Daniela
  Rus.
\newblock Lio-sam: Tightly-coupled lidar inertial odometry via smoothing and
  mapping.
\newblock In \emph{2020 IEEE/RSJ international conference on intelligent robots
  and systems (IROS)}, pages 5135--5142. IEEE, 2020.

\bibitem[Sun et~al.(2021)Sun, Xie, Chen, Zhou, and Bao]{sun2021neuralrecon}
Jiaming Sun, Yiming Xie, Linghao Chen, Xiaowei Zhou, and Hujun Bao.
\newblock Neuralrecon: Real-time coherent 3d reconstruction from monocular
  video.
\newblock In \emph{Proceedings of the IEEE/CVF Conference on Computer Vision
  and Pattern Recognition}, pages 15598--15607, 2021.

\bibitem[Tancik et~al.(2022)Tancik, Casser, Yan, Pradhan, Mildenhall,
  Srinivasan, Barron, and Kretzschmar]{tancik2022blocknerf}
Matthew Tancik, Vincent Casser, Xinchen Yan, Sabeek Pradhan, Ben Mildenhall,
  Pratul~P Srinivasan, Jonathan~T Barron, and Henrik Kretzschmar.
\newblock Block-nerf: Scalable large scene neural view synthesis.
\newblock In \emph{Proceedings of the IEEE/CVF Conference on Computer Vision
  and Pattern Recognition}, pages 8248--8258, 2022.

\bibitem[Tancik et~al.(2023)Tancik, Weber, Ng, Li, Yi, Wang, Kristoffersen,
  Austin, Salahi, Ahuja, et~al.]{tancik2023nerfstudio}
Matthew Tancik, Ethan Weber, Evonne Ng, Ruilong Li, Brent Yi, Terrance Wang,
  Alexander Kristoffersen, Jake Austin, Kamyar Salahi, Abhik Ahuja, et~al.
\newblock Nerfstudio: A modular framework for neural radiance field
  development.
\newblock In \emph{ACM SIGGRAPH 2023 Conference Proceedings}, pages 1--12,
  2023.

\bibitem[Tang et~al.(2019)Tang, Folkesson, and Jensfelt]{tang2019sparse2dense}
Jiexiong Tang, John Folkesson, and Patric Jensfelt.
\newblock Sparse2dense: From direct sparse odometry to dense 3-d
  reconstruction.
\newblock \emph{IEEE Robotics and Automation Letters}, 4\penalty0 (2):\penalty0
  530--537, 2019.

\bibitem[Tang et~al.(2020)Tang, Tian, Feng, Li, and Tan]{guidenetRGBLidar}
Jie Tang, Fei-Peng Tian, Wei Feng, Jian Li, and Ping Tan.
\newblock Learning guided convolutional network for depth completion.
\newblock \emph{IEEE Transactions on Image Processing}, 30:\penalty0
  1116--1129, 2020.

\bibitem[Tao et~al.(2023)Tao, Gao, Wang, Chen, Hao, Liang, Salzmann, and
  Yu]{tao2023lidarnerf}
Tang Tao, Longfei Gao, Guangrun Wang, Peng Chen, Dayang Hao, Xiaodan Liang,
  Mathieu Salzmann, and Kaicheng Yu.
\newblock Lidar-nerf: Novel lidar view synthesis via neural radiance fields.
\newblock \emph{arXiv preprint arXiv:2304.10406}, 2023.

\bibitem[Tosi et~al.(2023)Tosi, Tonioni, De~Gregorio, and
  Poggi]{tosi2023nerf4depth}
Fabio Tosi, Alessio Tonioni, Daniele De~Gregorio, and Matteo Poggi.
\newblock Nerf-supervised deep stereo.
\newblock In \emph{Proceedings of the IEEE/CVF Conference on Computer Vision
  and Pattern Recognition}, pages 855--866, 2023.

\bibitem[Turki et~al.(2023)Turki, Zhang, Ferroni, and Ramanan]{turki2023suds}
Haithem Turki, Jason~Y Zhang, Francesco Ferroni, and Deva Ramanan.
\newblock Suds: Scalable urban dynamic scenes.
\newblock In \emph{Computer Vision and Pattern Recognition (CVPR)}, 2023.

\bibitem[Uhrig et~al.(2017)Uhrig, Schneider, Schneider, Franke, Brox, and
  Geiger]{Uhrig2017THREEDV}
Jonas Uhrig, Nick Schneider, Lukas Schneider, Uwe Franke, Thomas Brox, and
  Andreas Geiger.
\newblock Sparsity invariant cnns.
\newblock In \emph{International Conference on 3D Vision (3DV)}, 2017.

\bibitem[Verbin et~al.(2022)Verbin, Hedman, Mildenhall, Zickler, Barron, and
  Srinivasan]{verbin2022refnerf}
Dor Verbin, Peter Hedman, Ben Mildenhall, Todd Zickler, Jonathan~T Barron, and
  Pratul~P Srinivasan.
\newblock Ref-nerf: Structured view-dependent appearance for neural radiance
  fields.
\newblock In \emph{2022 IEEE/CVF Conference on Computer Vision and Pattern
  Recognition (CVPR)}, pages 5481--5490. IEEE, 2022.

\bibitem[Walia et~al.(2022)Walia, Walz, Bijelic, Mannan, Julca-Aguilar, Langer,
  Ritter, and Heide]{gated2gated}
Amanpreet Walia, Stefanie Walz, Mario Bijelic, Fahim Mannan, Frank
  Julca-Aguilar, Michael Langer, Werner Ritter, and Felix Heide.
\newblock Gated2gated: Self-supervised depth estimation from gated images.
\newblock 2022.

\bibitem[Walz et~al.(2023)Walz, Bijelic, Ramazzina, Walia, Mannan, and
  Heide]{gatedstereo}
Stefanie Walz, Mario Bijelic, Andrea Ramazzina, Amanpreet Walia, Fahim Mannan,
  and Felix Heide.
\newblock Gated stereo: Joint depth estimation from gated and wide-baseline
  active stereo cues.
\newblock In \emph{Proceedings of the IEEE/CVF Conference on Computer Vision
  and Pattern Recognition}, pages 13252--13262, 2023.

\bibitem[Wang et~al.(2023{\natexlab{a}})Wang, Xu, Xu, and
  Lau]{wang2023lightingnerf}
Haoyuan Wang, Xiaogang Xu, Ke Xu, and Rynson~WH Lau.
\newblock Lighting up nerf via unsupervised decomposition and enhancement.
\newblock In \emph{Proceedings of the IEEE/CVF International Conference on
  Computer Vision}, pages 12632--12641, 2023{\natexlab{a}}.

\bibitem[Wang et~al.(2021)Wang, Liu, Liu, Theobalt, Komura, and
  Wang]{wang2021neus}
Peng Wang, Lingjie Liu, Yuan Liu, Christian Theobalt, Taku Komura, and Wenping
  Wang.
\newblock Neus: Learning neural implicit surfaces by volume rendering for
  multi-view reconstruction.
\newblock \emph{arXiv preprint arXiv:2106.10689}, 2021.

\bibitem[Wang et~al.(2023{\natexlab{b}})Wang, Shen, Gao, Huang, Munkberg,
  Hasselgren, Gojcic, Chen, and Fidler]{wang2023fegr}
Zian Wang, Tianchang Shen, Jun Gao, Shengyu Huang, Jacob Munkberg, Jon
  Hasselgren, Zan Gojcic, Wenzheng Chen, and Sanja Fidler.
\newblock Neural fields meet explicit geometric representations for inverse
  rendering of urban scenes.
\newblock In \emph{The IEEE Conference on Computer Vision and Pattern
  Recognition (CVPR)}, 2023{\natexlab{b}}.

\bibitem[Wong and Soatto(2021)]{wong2021unsupervised}
Alex Wong and Stefano Soatto.
\newblock Unsupervised depth completion with calibrated backprojection layers.
\newblock In \emph{Proceedings of the IEEE/CVF International Conference on
  Computer Vision}, pages 12747--12756, 2021.

\bibitem[Yang et~al.(2019)Yang, Manela, Happold, and Ramanan]{yang2019hsm}
Gengshan Yang, Joshua Manela, Michael Happold, and Deva Ramanan.
\newblock Hierarchical deep stereo matching on high-resolution images.
\newblock In \emph{The IEEE Conference on Computer Vision and Pattern
  Recognition (CVPR)}, 2019.

\bibitem[Yang et~al.(2017)Yang, Gao, and Shen]{yang2017real}
Zhenfei Yang, Fei Gao, and Shaojie Shen.
\newblock Real-time monocular dense mapping on aerial robots using
  visual-inertial fusion.
\newblock In \emph{2017 IEEE International Conference on Robotics and
  Automation (ICRA)}, pages 4552--4559. IEEE, 2017.

\bibitem[Yang et~al.(2023)Yang, Chen, Wang, Manivasagam, Ma, Yang, and
  Urtasun]{yang2023unisim}
Ze Yang, Yun Chen, Jingkang Wang, Sivabalan Manivasagam, Wei-Chiu Ma,
  Anqi~Joyce Yang, and Raquel Urtasun.
\newblock Unisim: A neural closed-loop sensor simulator.
\newblock In \emph{Proceedings of the IEEE/CVF Conference on Computer Vision
  and Pattern Recognition}, pages 1389--1399, 2023.

\bibitem[Yu et~al.(2021)Yu, Li, Tancik, Li, Ng, and
  Kanazawa]{yu2021plenoctrees}
Alex Yu, Ruilong Li, Matthew Tancik, Hao Li, Ren Ng, and Angjoo Kanazawa.
\newblock Plenoctrees for real-time rendering of neural radiance fields.
\newblock In \emph{Proceedings of the IEEE/CVF International Conference on
  Computer Vision}, pages 5752--5761, 2021.

\bibitem[Zhang et~al.(2023)Zhang, Zhang, Kuang, and Zhang]{zhang2023nerflidar}
Junge Zhang, Feihu Zhang, Shaochen Kuang, and Li Zhang.
\newblock Nerf-lidar: Generating realistic lidar point clouds with neural
  radiance fields.
\newblock \emph{arXiv preprint arXiv:2304.14811}, 2023.

\bibitem[Zhang et~al.(2020)Zhang, Riegler, Snavely, and
  Koltun]{zhang2020nerf++}
Kai Zhang, Gernot Riegler, Noah Snavely, and Vladlen Koltun.
\newblock Nerf++: Analyzing and improving neural radiance fields.
\newblock \emph{arXiv preprint arXiv:2010.07492}, 2020.

\bibitem[Zhang et~al.(2021)Zhang, Srinivasan, Deng, Debevec, Freeman, and
  Barron]{zhang2021nerfactor}
Xiuming Zhang, Pratul~P Srinivasan, Boyang Deng, Paul Debevec, William~T
  Freeman, and Jonathan~T Barron.
\newblock Nerfactor: Neural factorization of shape and reflectance under an
  unknown illumination.
\newblock \emph{ACM Transactions on Graphics (ToG)}, 40\penalty0 (6):\penalty0
  1--18, 2021.

\bibitem[Zhang et~al.(2022)Zhang, Wang, Li, Fu, and Guo]{SLFNetStereoLidar}
Yongjian Zhang, Longguang Wang, Kunhong Li, Zhiheng Fu, and Yulan Guo.
\newblock Slfnet: A stereo and lidar fusion network for depth completion.
\newblock \emph{IEEE Robotics and Automation Letters}, 7\penalty0 (4):\penalty0
  10605--10612, 2022.

\bibitem[Zhou et~al.(2017)Zhou, Brown, Snavely, and Lowe]{Zhou2017}
Tinghui Zhou, Matthew Brown, Noah Snavely, and David~G. Lowe.
\newblock Unsupervised learning of depth and ego-motion from video.
\newblock In \emph{CVPR}, 2017.

\end{thebibliography}
}

\end{document}